\documentclass[letterpaper]{article} 
\usepackage{aaai2026}  
\usepackage{times}  
\usepackage{helvet}  
\usepackage{courier}  
\usepackage[hyphens]{url}  
\usepackage{graphicx} 
\usepackage{natbib}  
\usepackage{caption} 
\usepackage{algorithm}
\usepackage{algorithmic}
\usepackage{subcaption}
\usepackage{newfloat}
\usepackage{listings}
\usepackage{booktabs}
\usepackage{multirow}
\usepackage{amsmath, amsthm}
\usepackage{amsfonts} 
\usepackage{amssymb}
\newtheorem{assumption}{Assumption}[section]
\newtheorem{lemma}{Lemma}[section]
\newtheorem{theorem}{Theorem}
\DeclareCaptionStyle{ruled}{labelfont=normalfont,labelsep=colon,strut=off} 
\floatstyle{ruled}
\newfloat{listing}{tb}{lst}{}
\floatname{listing}{Listing}
\title{HeteroTune: Efficient  Federated Learning for Large Heterogeneous Models}
\author{
	Ruofan Jia\textsuperscript{\rm 1}, 
	Weiying Xie\textsuperscript{\rm 1}, 
	Jie Lei\textsuperscript{\rm 2}, 
	Jitao Ma\textsuperscript{\rm 1}, 
	Haonan Qin\textsuperscript{\rm 1}, 
	Leyuan Fang\textsuperscript{\rm 3}
}
\affiliations{
	\textsuperscript{\rm 1}Xidian University, 
	\textsuperscript{\rm 2}University of Technology Sydney, 
	\textsuperscript{\rm 3}Hunan University
}

\usepackage{bibentry}

\begin{document}

\maketitle

\begin{abstract}
While large pre-trained models have achieved impressive performance across AI tasks, their deployment in privacy-sensitive and distributed environments remains challenging. Federated learning (FL) offers a viable solution by enabling decentralized fine-tuning without data sharing, but real-world applications face significant obstacles due to heterogeneous client resources in compute and memory.
To address this, we propose HeteroTune, a novel federated fine-tuning paradigm for large, heterogeneous models operating under limited communication and computation budgets.
The core of our method lies in a novel architecture, DeMA (Dense Mixture of Adapters), which enables flexible and efficient aggregation of heterogeneous models by preserving their full representational capacity while facilitating seamless cross-model knowledge fusion.
We further introduce CMGA (Cross-Model Gradient Alignment), a lightweight yet effective mechanism that enhances training stability by harmonizing gradient directions across heterogeneous client models during aggregation, mitigating update conflicts and promoting more consistent convergence in federated settings.
We provide both \textit{theoretical} analysis and \textit{empirical} evidence showing that HeteroTune achieves state-of-the-art performance and efficiency across diverse tasks and model architectures. For example, on LLaMA models, it reduces communication overhead by 99.5\%, cuts peak memory usage by ~50\%, and improves performance by 4.61\%.
\end{abstract}


\section{Introduction}

The rise of open-source large-scale pre-trained Transformer models \cite{attention, pre, pre1, llama, gemma, deepseek} has not only driven breakthroughs in AI applications, but also sparked a growing demand for their deployment and fine-tuning.
However, privacy constraints and widespread data silos pose major challenges for fine-tuning large models \cite{island, island1}. 
Federated learning (FL) \cite{fedavg}, as a privacy-preserving distributed learning paradigm, provides a promising solution for enabling large model collaboration at the edge. 
Even so, the real-world deployment of FL is significantly challenged by resource heterogeneity, primarily in terms of computational power and memory capacity. To be specific, excluding less capable clients results in insufficient data utilization, while scaling down model sizes to enable broader participation often compromises the representing capacity  \cite{inclusivefl, fiarse}.

To address this issue, a practical solution is to adopt model-heterogeneity federated learning (MHFL), where the model on each client is tailored to match its computational capabilities. Existing approaches can be broadly categorized into three types based on their implementation strategies: knowledge distillation-based methods \cite{fedmd, dfrd, feddfa, fedgkd}, sub-model extraction methods \cite{fiarse, flexifed, fedlmt, heterofl, scalefl, nefl}, and proxy-based methods \cite{bridging, mh-pflid, pfedes, moss}, as illustrated in Figure \ref{fig1a}, \ref{fig1b} and \ref{fig1c}. Among these approaches:
(a) Knowledge distillation-based methods aggregate knowledge using soft labels predicted by local models. However, they typically require access to additional public datasets and incur substantial computational overhead related to the distillation process, which hinders their practical deployment.
(b) Sub-model extraction methods flexibly adapt to client capabilities by extracting appropriately sized sub-models and performing global adaptive aggregation. While effective in certain scenarios, this approach is less suitable for fine-tuning large pre-trained models, as sub-model extraction may disrupt the pre-trained architecture of the model, leading to convergence difficulties and degraded generalization performance.
(c) Proxy-based methods introduce a proxy model to facilitate knowledge transfer across heterogeneous clients. However, when the size gap between local models and the proxy becomes too large, injecting knowledge into the proxy becomes highly challenging. Moreover, such methods often incur additional computational overhead due to the training of the proxy model.

\begin{figure*}[htbp]
	\centering
	\includegraphics[width=\textwidth]{abs.pdf}
	\begin{minipage}[b]{0.24\textwidth}
		\phantomsubcaption
		\label{fig1a}
	\end{minipage}
	\begin{minipage}[b]{0.24\textwidth}
		\phantomsubcaption
		\label{fig1b}
	\end{minipage}
	\begin{minipage}[b]{0.24\textwidth}
		\phantomsubcaption
		\label{fig1c}
	\end{minipage}
	\begin{minipage}[b]{0.24\textwidth}
		\phantomsubcaption
		\label{fig1d}
	\end{minipage}
	\caption{Comparison of HeteroTune with existing model-heterogeneous federated learning (MHFL) approaches.
		(a) Distillation-based methods require public data and introduce extra computation.
		(b) Submodel extraction disrupts model structure, reducing representational capacity.
		(c) Proxy-based methods struggle with large model gaps and added overhead.
		(d) In contrast, HeteroTune leverages DeMA to enable efficient, structure-preserving knowledge sharing across heterogeneous models without relying on external data or incurring heavy computation.
	}
	\label{fig1}
\end{figure*}
To overcome the limitations of the above methods, we propose \textbf{HeteroTune}, an adapter-based \cite{peft} federated fine-tuning framework that enables efficient knowledge sharing across heterogeneous models.
Inspired by the theory of knowledge decomposition in federated learning \cite{decoupling, cho, gpfl}, we decompose traditional adapters into a pair of modules: a local-adapter and a share-adapter, and introduce the \underline{De}nse \underline{M}ixed of \underline{A}dapter (\textbf{DeMA}) architecture to support knowledge aggregation across models with varying scales.
DeMA employs densely connected aggregation of adapters, allowing comprehensive retention and flexible fusion of knowledge across heterogeneous models (see Figure \ref{fig1d}). This design mitigates knowledge entanglement issues inherent in sub-model extraction approaches, while also avoiding the additional overhead commonly associated with distillation-based methods. Furthermore, to improve inference efficiency, DeMA merges the parameter matrices of the adapter pair during inference by absorbing the share-adapter into the local-adapter. This transformation is mathematically equivalent to the original dual-adapter formulation, ensuring no loss in representational capacity while reducing computational overhead at inference time. 
Additionally, to address the optimization instability caused by conflicting updates from heterogeneous clients, we introduce Cross-Model Gradient Alignment (\textbf{CMGA}), a gradient-level coordination mechanism that fine-tunes the aggregation of shared model components. CMGA minimizes interference between divergent optimization directions, enabling stable convergence toward an optimal global model. This enhances both the accuracy and efficiency of federated learning across diverse client models.
Finally, extensive experiments in both computer vision (CV) and natural language processing (NLP) domains demonstrate that our HeteroTune outperforms existing methods in terms of performance, while maintaining high efficiency in computation and communication. 
Our main contributions include the following:
\begin{itemize}
	\setlength{\leftmargin}{-2pt}
	\item \textbf{Algorithmically}, we propose HeteroTune, a federated fine-tuning framework for heterogeneous large models. At its core is DeMA, a densely connected adapter architecture for effective cross-model knowledge fusion, paired with CMGA, a gradient alignment mechanism that stabilizes optimization across clients.
	\item \textbf{Theoretically}, we establish the convergence guarantees of HeteroTune by analyzing its adapter-based optimization dynamics within a federated setting, showing that the framework converges under non-convex objectives and heterogeneous data distributions commonly encountered in practical scenarios.
	\item \textbf{Empirically}, we conduct extensive experiments on both CV and  NLP tasks, demonstrating that HeteroTune consistently outperforms existing baselines. The results validate the effectiveness and generalizability of our approach across a wide range of large-scale model architectures and downstream applications.
\end{itemize}

\begin{figure*}[ht]
	\begin{minipage}[b]{0.5\textwidth}
		\phantomsubcaption
		\label{fig2a}
	\end{minipage}
	\begin{minipage}[b]{0.5\textwidth}
		\phantomsubcaption
		\label{fig2b}
	\end{minipage}
	\centering
	\includegraphics[width=0.9\textwidth]{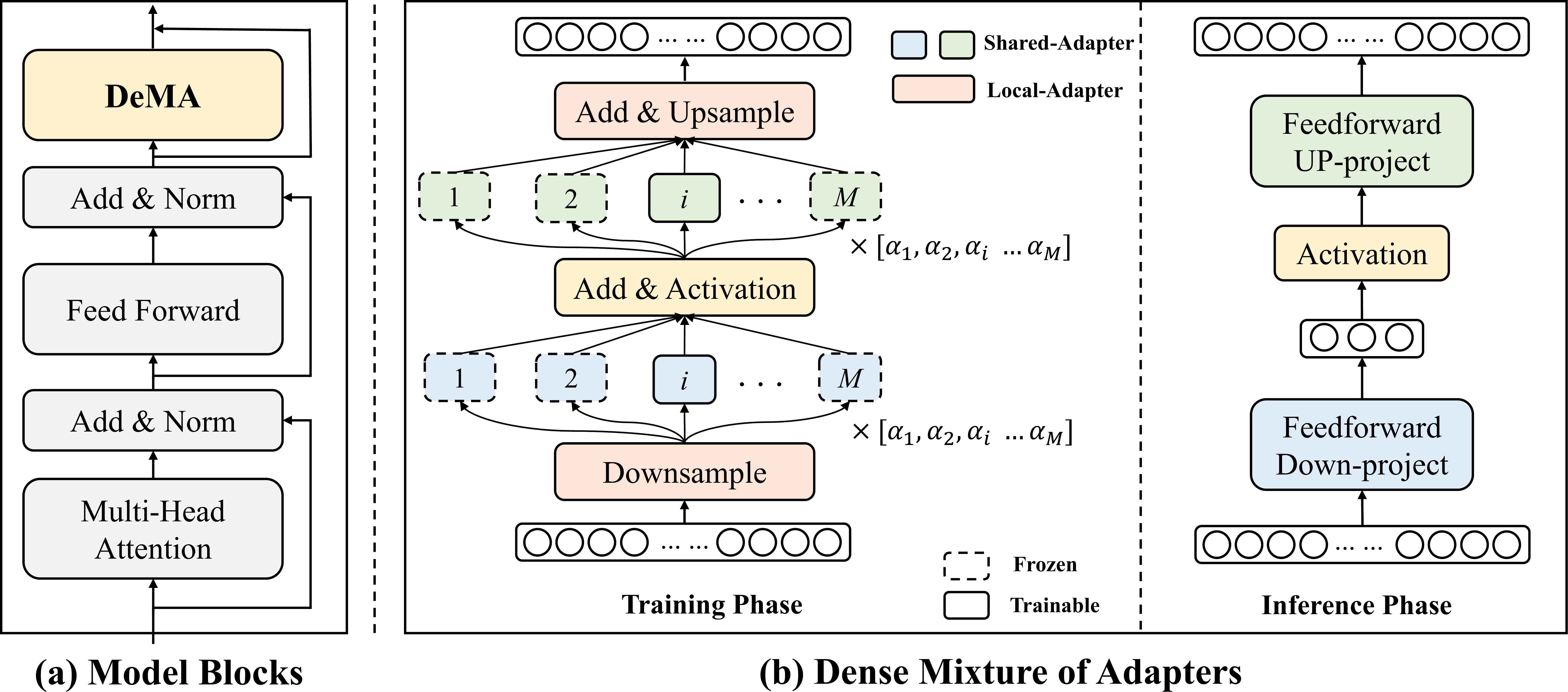}
	\caption{
		(a) DeMA is integrated into each Transformer block to enable modular and efficient fine-tuning.
		(b) The internal structure includes a local adapter for projection and multiple densely connected share-adapters combined via learnable weights $[\alpha_1, \alpha_2, \dots, \alpha_M]$. During training, only the $i$-th share-adapter is updated for each client, while others remain frozen. At inference, all adapters are merged into a single linear layer for efficiency.}
\end{figure*}
\section{Related Work}
\subsection{Model-Heterogeneity Federated Learning}
MHFL addresses client resource heterogeneity by aggregating knowledge from heterogeneous models. HeteroFL \cite{heterofl} achieves global model aggregation across differently sized local models by aligning and merging overlapping parameters. InclusiveFL \cite{inclusivefl} addresses parameter mismatch through layer-wise aggregation, while ScaleFL \cite{scalefl} adaptively scales networks along width and depth dimensions using early exits. FIARSE \cite{fiarse} dynamically adjusts submodels based on parameter importance. However, these methods require full model training and cannot leverage pre-trained weights for transfer learning.
MH-pFLID \cite{mh-pflid} proposes a lightweight messenger model for knowledge transfer, and Moss \cite{moss} achieves full-weight aggregation through proxy models. These approaches face challenges in capturing large model representations while incurring substantial computational overhead. Knowledge distillation frameworks \cite{feddf,fedmd,dfrd} enable heterogeneous model aggregation but introduce additional computational costs and require public datasets.

\subsection{PEFT for Federated Learning}
To reduce communication and computational costs, several studies integrate PEFT techniques into federated learning. Works such as \cite{fedtune, oscar, fedclip} fine-tune CLIP \cite{clip} models by updating limited learnable parameters. Yang et al. \cite{diff} explores diffusion \cite{diffusion} model fine-tuning for generative FL tasks. FedDAT \cite{feddat} employs adapter strategies for multimodal foundation models while addressing data heterogeneity.
Some studies apply PEFT for MHFL. FedLMT \cite{fedlmt} tackles system heterogeneity via low-rank model training but alters model architecture, potentially causing parameter instability. FLoRA \cite{flora} leverages LoRA for efficient MHFL with stacking-based aggregation for noise-free parameter fusion. However, this approach lacks semantic disentanglement between heterogeneous updates, potentially causing knowledge entanglement and undermining representation fidelity.
\section{Method}
Our goal is to enable efficient federated learning across clients with heterogeneous model capacities. We achieve this by allowing clients to train models of different scales based on their resource constraints, and designing an aggregation strategy that effectively fuses their updates. 
We first introduce DeMA, a new architecture for efficient fine-tuning in heterogeneous federated learning. Building upon this foundation, we then introduce CMGA, a key algorithmic innovation that effectively resolves gradient conflicts arising between heterogeneous models.  Finally, we summarize the overall framework of HeteroTune.

\subsection{DeMA}
\textbf{Adapter.} An adapter typically integrates a down-projection matrix, an up-projection matrix, and an intermediate nonlinear activation operation within the original model. This structure provides a lightweight approach to incorporating new features and knowledge while keeping most of the model parameters frozen, enabling efficient fine-tuning and adaptive extension.
For an input $x\in \mathbb{R}^{n \times m}$, the adapters are integrated with a residual connection, resulting in the final transformation:
\begin{equation}
	x \rightarrow x+h(xA)B,
\end{equation}
where $h(\cdot)$ is a nonlinear activation function, and $A\in\mathbb{R}^{m\times r}$ and $B\in \mathbb{R}^{r\times m}$ represent the up-projection matrix and down-projection matrix, respectively.

\textbf{Architecture.} To enable effective aggregation in heterogeneous model settings, we propose a novel adapter-based architecture named DeMA, which is inserted into each block of models (see Figure \ref{fig2a}). DeMA decouples adaptation into a local-adapter and a set of share-adapters. The local-adapter is uniquely tailored to each type of model, playing a similar role to traditional adapters by enabling coarse-grained, model-specific adjustments. 
It consists of a downsampling matrix and an upsampling matrix, which are denoted by $Wl_{down}$ and $Wl_{up}$ respectively.
In addition, each DeMA module contains $M$ square share-adapter matrices, where $M$ corresponds to the number of distinct model types participating in FL. Each model variant is associated with a distinct share-adapter, enabling model-specific knowledge contribution.
Similar to the local-adapter, two sets of share-adapters are positioned before and after the non-linear activation function. 
As before, we denote these two sets of parameters by $Ws_{down}$ and $Ws_{up}$, respectively.
Building upon the local-adapter’s projection into a common latent space, the share-adapter enables fine-grained knowledge aggregation across heterogeneous models, allowing DeMA to simultaneously preserve model individuality and facilitate collaborative learning.

\textbf{Training and inference.}
During training, the local-adapter and the share-adapter corresponding to its own model are trainable, while the other 
$M-1$ share-adapters corresponding to different models are frozen (see Figure \ref{fig2b}).
In addition, we introduce a set of learnable gating parameters, denoted as $\alpha_i$, that remain active during training, enabling adaptive optimization of the share-adapters' contributions.
This densely connected multi-branch architecture ensures that all models, despite structural differences, can contribute to and benefit from shared representation learning in a federated setting.
In a word, the transformation of training phase is:
\begin{equation}
	x \rightarrow x + h(x Wl_{down}\sum_{i=1}^M Ws_{down}^i) \sum_{i=1}^M Ws_{up}^i Wl_{up}.
\end{equation}
To maintain efficiency and scalability during inference, DeMA adopts a parameter merging strategy. Specifically, the share-adapters are absorbed into the local-adapter through a mathematically equivalent transformation, resulting in a single feedforward adapter block (see Figure \ref{fig2b}). This design eliminates the need for runtime multi-branch computation, significantly reducing the inference overhead while preserving the knowledge integration achieved during training. As such, DeMA balances the richness of multi-source learning with the efficiency of a lightweight inference path, making it well-suited for deployment in real-world, resource-constrained federated environments.
Thus, we can easily derive the simplified formula:
\begin{equation}
	\begin{gathered}
		A' = Wl_{down}\sum_{i=1}^M Ws_{down}^i,
		B' = \sum_{i=1}^M Ws_{up}^i Wl_{up},\\
		x \rightarrow x + h(xA')B'.
	\end{gathered}
\end{equation}
\subsection{CMGA}
In HeteroTune, the share-adapter is introduced to facilitate knowledge transfer across heterogeneous models. However, when aggregating gradients from structurally diverse clients, the share-adapter updates can exhibit high variance in both direction and scale, leading to instability and suboptimal convergence. To mitigate these issues, we propose \textbf{Cross-Model Gradient Alignment (CMGA)}, a lightweight alignment mechanism that regularizes the update geometry across clients.
Let $\nabla Ws_i \in \mathbb{R}^d$ represent the gradient of the share-adapter associated with the $i^{th}$ type of model. We stack these gradients into a matrix:
\begin{equation}
	G_s = [\nabla Ws_1 \mid \nabla Ws_2 \mid \cdots \mid \nabla Ws_M] \in \mathbb{R}^{d \times M},
\end{equation}
which characterizes the geometric configuration of the share-adapter updates. A key observation is that the more orthogonal and balanced these gradient directions are, the more stable the aggregated update becomes. Therefore, we seek a transformed version $\widetilde{G}_s$ that preserves the information in $G_s$ while improving its alignment properties.
Specifically, CMGA solves the following constrained  problem:
\begin{equation}
	\widetilde{G}_s = \arg\min_{Z \in \mathbb{R}^{d \times N}} \|Z - G_s\|_F^2 \quad \text{s.t.} \quad \text{Cov}(Z) = I,
\end{equation}
where $\text{Cov}(Z) = \frac{1}{N} Z Z^\top$ denotes the covariance structure of the aligned gradients. The identity constraint ensures that all directions contribute equally and no single client dominates the update.
This problem admits a closed-form solution using whitening and rescaling:

\begin{equation}
	\widetilde{G}_s = G_s H, \quad \text{with} \quad H = (G_s^\top G_s + \epsilon I)^{-\frac{1}{2}},
\end{equation}
where $H \in \mathbb{R}^{M \times M}$ serves as an alignment kernel and $\epsilon$ is a small positive constant for numerical stability. The aligned gradient matrix $\widetilde{G}_s$ is then aggregated and applied to update the shared parameters.
This approach introduces no additional communication, has negligible computation cost on the server, and significantly reduces gradient interference across clients. Empirically, we observe that CMGA leads to faster convergence and more stable performance.
\begin{figure}
	\includegraphics[width=0.45\textwidth]{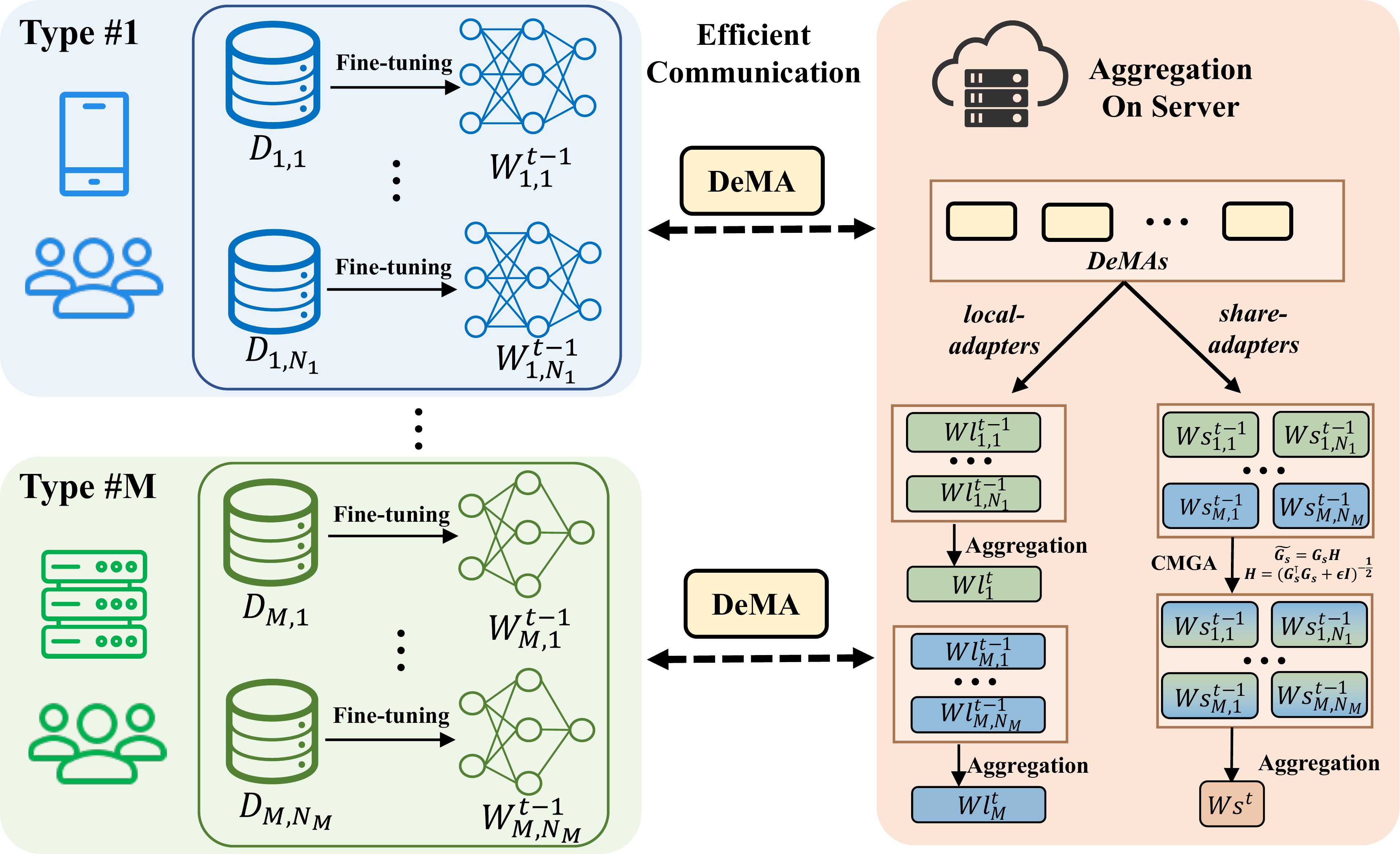}
	\caption{Illustration of the HeteroTune framework. Clients are grouped by type and perform local fine-tuning on their data. On the server, local-adapters are aggregated by type, while share-adapters are first processed by CMGA to resolve gradient conflicts, then aggregated using FedAvg. Only DeMAs are transmitted between clients and the server, enabling efficient communication.}
	\label{framework}
	\vspace{-4mm}
\end{figure}
\begin{table*}[h]
	\centering
	\setlength{\tabcolsep}{4pt}
	\begin{tabular}{c|ccccc|ccccc}
		\toprule
		\multirow{2}{*}{Method} 
		& \multicolumn{5}{c|}{\textbf{CIFAR-10}} 
		& \multicolumn{5}{c}{\textbf{CIFAR-100}} \\
		& Tiny & Small & Base & Large & Avg 
		& Tiny & Small & Base & Large & Avg \\
		\midrule
		AllSmall      & 62.25\textsubscript{±0.19} & -- & -- & -- & 62.25 & 54.77\textsubscript{±0.14} & -- & -- & -- & 54.77 \\
		AllLarge      & -- & -- & -- & 80.76\textsubscript{±0.21} & 80.76 & -- & -- & -- & 62.27\textsubscript{±0.27} & 62.27 \\
		Homo-Training & 60.64\textsubscript{±0.26} & 62.51\textsubscript{±0.18} & 67.11\textsubscript{±0.22} & 74.29\textsubscript{±0.25} & 66.14 & 50.88\textsubscript{±0.17} & 52.50\textsubscript{±0.14} & 55.23\textsubscript{±0.23} & 58.43\textsubscript{±0.16} & 54.26 \\
		FedDF         & 67.42\textsubscript{±0.11} & 67.89\textsubscript{±0.26} & 69.53\textsubscript{±0.15} & 74.69\textsubscript{±0.19} & 69.88 & 54.81\textsubscript{±0.13} & 55.10\textsubscript{±0.21} & 58.03\textsubscript{±0.20} & 60.70\textsubscript{±0.22} & 57.16 \\
		ScaleFL       & 52.85\textsubscript{±0.24} & 61.24\textsubscript{±0.29} & 65.49\textsubscript{±0.18} & 69.32\textsubscript{±0.17} & 62.22 & 49.34\textsubscript{±0.16} & 51.42\textsubscript{±0.14} & 55.56\textsubscript{±0.28} & 55.72\textsubscript{±0.27} & 53.01 \\
		MH-pFLID      & 65.54\textsubscript{±0.27} & 68.37\textsubscript{±0.23} & 71.86\textsubscript{±0.25} & 74.79\textsubscript{±0.13} & 70.14 & 55.72\textsubscript{±0.29} & 58.12\textsubscript{±0.20} & 59.75\textsubscript{±0.17} & 60.65\textsubscript{±0.19} & 58.56 \\
		FLoRA         & 64.53\textsubscript{±0.18} & 68.37\textsubscript{±0.15} & 73.49\textsubscript{±0.22} & 77.11\textsubscript{±0.28} & 70.78 & 54.27\textsubscript{±0.24} & 55.92\textsubscript{±0.12} & 59.81\textsubscript{±0.19} & 60.36\textsubscript{±0.14} & 57.59 \\
		\textbf{HeteroTune} & \textbf{66.91\textsubscript{±0.14}} & \textbf{71.50\textsubscript{±0.11}} & \textbf{75.66\textsubscript{±0.25}} & \textbf{80.95\textsubscript{±0.16}} & \textbf{73.75} & \textbf{57.42\textsubscript{±0.26}} & \textbf{59.97\textsubscript{±0.18}} & \textbf{62.04\textsubscript{±0.22}} & \textbf{62.79\textsubscript{±0.19}} & \textbf{60.55} \\
		\midrule
		\multirow{2}{*}{Method} 
		& \multicolumn{5}{c|}{\textbf{Caltech-101}} 
		& \multicolumn{5}{c}{\textbf{Caltech-256}} \\
		& Tiny & Small & Base & Large & Avg 
		& Tiny & Small & Base & Large & Avg \\
		\midrule
		AllSmall      & 73.85\textsubscript{±0.13} & -- & -- & -- & 73.85 & 59.81\textsubscript{±0.26} & -- & -- & -- & 59.81 \\
		AllLarge      & -- & -- & -- & 88.74\textsubscript{±0.18} & 88.74 & -- & -- & -- & 68.85\textsubscript{±0.14} & 68.85 \\
		Homo-Training & 70.42\textsubscript{±0.27} & 72.61\textsubscript{±0.19} & 76.86\textsubscript{±0.17} & 79.31\textsubscript{±0.28} & 74.80 & 55.37\textsubscript{±0.13} & 56.46\textsubscript{±0.22} & 57.20\textsubscript{±0.23} & 62.13\textsubscript{±0.29} & 57.79 \\
		FedDF         & 72.03\textsubscript{±0.26} & 72.58\textsubscript{±0.12} & 75.99\textsubscript{±0.25} & 79.55\textsubscript{±0.23} & 75.11 & 60.56\textsubscript{±0.16} & 62.02\textsubscript{±0.29} & 63.09\textsubscript{±0.14} & 63.61\textsubscript{±0.24} & 62.32 \\
		ScaleFL       & 66.74\textsubscript{±0.17} & 67.73\textsubscript{±0.11} & 69.32\textsubscript{±0.21} & 72.06\textsubscript{±0.16} & 68.91 & 46.27\textsubscript{±0.18} & 47.71\textsubscript{±0.27} & 50.33\textsubscript{±0.13} & 50.97\textsubscript{±0.19} & 48.82 \\
		MH-pFLID      & 76.40\textsubscript{±0.25} & 79.25\textsubscript{±0.12} & 80.67\textsubscript{±0.29} & 81.56\textsubscript{±0.27} & 79.72 & 59.82\textsubscript{±0.15} & 61.22\textsubscript{±0.28} & 62.80\textsubscript{±0.23} & 63.88\textsubscript{±0.19} & 61.93 \\
		FLoRA         & 75.37\textsubscript{±0.19} & 78.35\textsubscript{±0.16} & 79.99\textsubscript{±0.28} & 80.81\textsubscript{±0.26} & 78.63 & 60.12\textsubscript{±0.14} & 61.24\textsubscript{±0.23} & 63.46\textsubscript{±0.17} & 64.46\textsubscript{±0.29} & 62.32 \\
		\textbf{HeteroTune} & \textbf{77.65\textsubscript{±0.21}} & \textbf{83.59\textsubscript{±0.22}} & \textbf{86.70\textsubscript{±0.24}} & \textbf{87.61\textsubscript{±0.16}} & \textbf{83.89} & \textbf{62.65\textsubscript{±0.17}} & \textbf{65.12\textsubscript{±0.29}} & \textbf{67.25\textsubscript{±0.26}} & \textbf{68.57\textsubscript{±0.18}} & \textbf{65.89} \\
		\bottomrule
	\end{tabular}
	
	\caption{The comparison results of our method with the baseline and existing methods using ViT series models on CV tasks under Dirichlet non-IID with $\alpha=0.1$. The performance of each model variant is reported, along with the averaged  results of all models.
		Our HeteroTune, highlighted in bold, achieves the best performance.}
	\label{table1}
\end{table*}
\subsection{HeteroTune}
Figure \ref{framework} outlines the training procedure of HeteroTune. 
At initialization, each client is assigned a model from one of $M$ predefined model types based on its resource capacity.
For ease of exposition, we assume there are $M$ distinct types of models. The clients associated with each model type form a group, and we denote by $N_i$ the number of clients in the $i^{th}$ group.
Furthermore, we use subscripts and superscripts to distinguish between these models and to indicate the iterative process across different rounds. 
Specifically, we denote by $W_{i,j}^t$ the model associated with the $j^{th}$ client in the $i^{th}$ group during the $t^{th}$ round of training, where $1\leq i\leq M$ and $1 \leq j \leq N_i$.

Each comunication round consists of two phases: client update and server aggregation. In the client update phase, the client receives the latest DeMAs from the server and performs local training on its local model. These DeMAs are then sent back to the server.
In the server update phase, HeteroTune performs a two-stage aggregation. First, a homogeneous aggregation groups clients by model type and applies FedAvg to their local-adapters, yielding updated $Wl_i^t$ for each model group. 
In $Wl_i^t$, the subscript $i$ refers to the $i^{th}$ group, and the superscript $t$ indicates the parameters after the $t^{th}$ training round.
Then, for cross-model knowledge sharing, all shared adapters are first processed by CMGA to address gradient conflicts. Subsequently, they are aggregated using FedAvg to produce $Ws^t$. Finally, each client's local model is updated using the newly aggregated pair $\{Wl_i^t, Ws^t\}$, and the next round begins.

\section{Convergence Analysis}

We present a rigorous, self‑contained convergence analysis of HeteroTune in the non‑convex federated setting under standard smoothness and bounded‑variance assumptions.  
Our main result establishes that the algorithm reaches an expected first-order stationarity of the global objective at the canonical rate
\(\mathcal{O}(1/\sqrt{T})\).  
To derive Theorem~1, we build on four key lemmas:

\begin{itemize}
	\item \textbf{Lemma 1:} Local SGD yields descent on each client: the objective decreases proportionally to the squared gradient norm, with an error term depending on variance and parameter drift.
	
	\item \textbf{Lemma 2:} The group-averaged local gradients are unbiased estimators of the true gradients, with variance reduced by the group size.
	
	\item \textbf{Lemma 3:} Whitening ensures that group gradients are decorrelated and have identity covariance, enabling tight control of representation gradient variance.
	
	\item \textbf{Lemma 4:} Each round yields expected descent in the global objective, balancing a negative gradient norm term and a controlled variance term with constants \(C_1 = d/2\), \(C_2 = \frac{1}{2}(1+E)\).
\end{itemize}
\textbf{Theorem~1 (Expected First-Order Stationarity).}
Let $W^{(t)} = (Ws^{(t)}\in\mathbb{R}^{r\times r} , \{Wl^{(t)}\in\mathbb{R}^{m\times r}\}_{g=1}^M)$ denote the parameters at round $t$ and $F_{\inf} = \inf_{W} F(W)$. 
For any total number of rounds $T \ge 1$, we have
\begin{align}
	\min_{0 \le t < T} 
	\mathbb{E}\Bigl[\,
	\|\nabla F(W^{(t)})\|^2
	\Bigr]
	&\;\le\;
	\frac{2\bigl(F(W^{(0)}) - F_{\inf}\bigr)} 
	{\eta_0 \sqrt{T}} \notag \\
	&\quad+
	\frac{2 \eta_0}{\sqrt{T}}
	\Bigl(
	C_1 \sigma^2 + C_2 E^2 L^2
	\Bigr),
\end{align}
where $\eta_0$ denotes  the initial server step size, $E$ is the number of local SGD steps, $L$ is the common $L$-smooth constant, and $\sigma^2$ is an upper bound on mini‑batch‑gradient variance. The constants in the bound satisfy $C_1 = \tfrac{r}{2},  C_2 = \tfrac{1}{2}(1+E)$.
In other words, HeteroTune achieves the standard
$\mathcal{O}(1/ \sqrt{T})$
rate toward an expected first-order stationary point in the non-convex setting.

\paragraph{Proof sketch.}
(i) Apply the smoothness descent lemma to the joint parameter vector $W$ after $E$ local SGD steps.  
(ii) Show that group-wise averaging is unbiased and reduces variance.  
(iii) Prove that whitening decorrelates the group-level gradients and caps their covariance by the identity matrix.  
(iv) Combine the above to obtain a per-round descent bound, sum over rounds using $\eta_t \propto 1/\sqrt{t}$, and take the minimal gradient norm.  
The full derivations and proofs of the supporting lemmas are deferred to the Appendix.

\begin{table*}[t]
	\small
	\centering
	\setlength{\tabcolsep}{3pt}
	\begin{tabular}{c|cccc|cccc|cccc}
		\toprule
		\multirow{2}{*}{Method} & \multicolumn{4}{c|}{\textbf{SST-2}} &
		\multicolumn{4}{c|}{\textbf{CoLA}} &
		\multicolumn{4}{c}{\textbf{QQP}} \\
		&1B&3B&8B&Avg&1B&3B&8B&Avg&1B&3B&8B&Avg\\
		\midrule
		AllSmall &92.16\textsubscript{±0.21}&--&--&92.16&79.42\textsubscript{±0.18}&--&--&79.42&88.50\textsubscript{±0.32}&--&--&88.50\\
		AllLarge&--&--&95.71\textsubscript{±0.27}&95.71&--&--&86.42\textsubscript{±0.43}&86.42&--&--&92.07\textsubscript{±0.31}&92.07\\
		Homo-Training&91.74\textsubscript{±0.35}&92.69\textsubscript{±0.42}&93.67\textsubscript{±0.19}&92.70&76.67\textsubscript{±0.29}&82.62\textsubscript{±0.34}&84.40\textsubscript{±0.22}&81.23&83.28\textsubscript{±0.48}&86.41\textsubscript{±0.30}&91.06\textsubscript{±0.21}&86.92\\
		FedDF&92.35\textsubscript{±0.17}&93.36\textsubscript{±0.39}&94.40\textsubscript{±0.24}&93.37&76.32\textsubscript{±0.26}&81.54\textsubscript{±0.31}&82.47\textsubscript{±0.41}&80.11&86.36\textsubscript{±0.36}&88.85\textsubscript{±0.23}&89.31\textsubscript{±0.19}&88.17\\
		ScaleFL&86.34\textsubscript{±0.44}&90.71\textsubscript{±0.36}&91.75\textsubscript{±0.28}&89.60&70.17\textsubscript{±0.40}&74.77\textsubscript{±0.21}&75.83\textsubscript{±0.43}&73.59&80.48\textsubscript{±0.39}&83.26\textsubscript{±0.33}&85.17\textsubscript{±0.37}&82.97\\
		MH-pFLID&92.59\textsubscript{±0.15}&91.14\textsubscript{±0.47}&93.50\textsubscript{±0.29}&92.41&74.30\textsubscript{±0.34}&78.59\textsubscript{±0.20}&80.57\textsubscript{±0.23}&77.82&85.41\textsubscript{±0.42}&87.73\textsubscript{±0.35}&88.70\textsubscript{±0.41}&87.28\\
		FLoRA&92.22\textsubscript{±0.24}&93.18\textsubscript{±0.28}&93.57\textsubscript{±0.26}&92.99&76.68\textsubscript{±0.41}&81.69\textsubscript{±0.31}&82.59\textsubscript{±0.39}&80.32&88.34\textsubscript{±0.36}&87.15\textsubscript{±0.28}&90.10\textsubscript{±0.35}&88.53\\
		\textbf{HeteroTune} &\textbf{93.95\textsubscript{±0.22}}&\textbf{94.01\textsubscript{±0.19}}&\textbf{94.67\textsubscript{±0.21}}&
		\textbf{94.21}&\textbf{79.83\textsubscript{±0.34}}&\textbf{83.50\textsubscript{±0.31}}&\textbf{84.41\textsubscript{±0.27}}&\textbf{82.58}&\textbf{89.76\textsubscript{±0.26}}&\textbf{91.07\textsubscript{±0.18}}&\textbf{92.11\textsubscript{±0.22}}&\textbf{90.98}\\
		\bottomrule
	\end{tabular}
	\vspace{-2mm}
	\caption{The comparison results of our method with the baseline and existing methods using LLaMA series models on GLUE datasets for NLP tasks. 
		The performance of each model variant is reported, along with the averaged accuracy results of all models.
		Our HeteroTune, highlighted in bold, achieves the best performance.}
	\label{table2}
\end{table*}
\begin{figure*}[t]
	\centering
	\includegraphics[width=\textwidth]{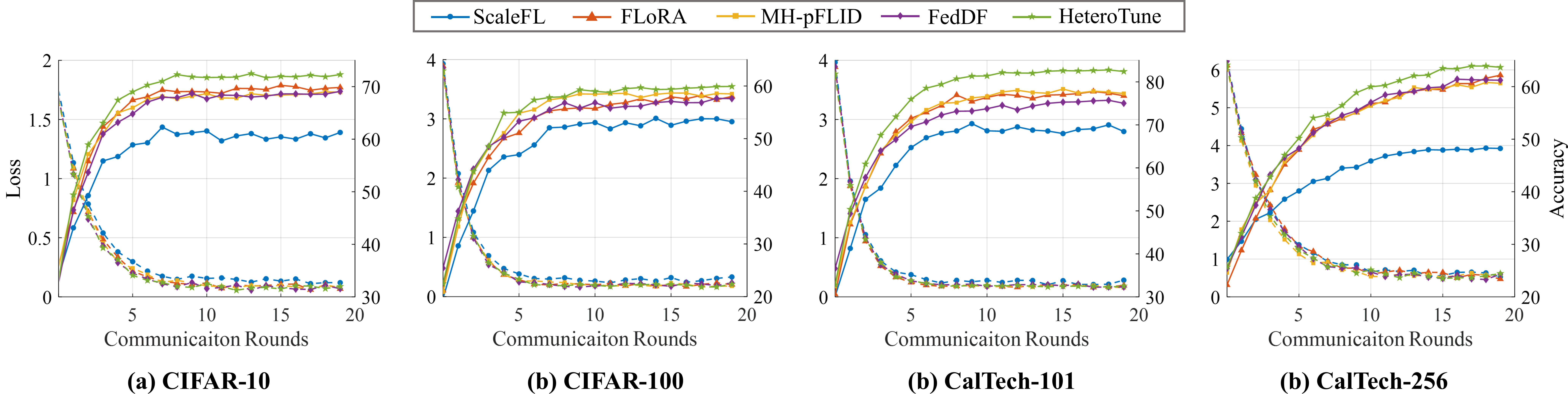}
	\caption{Convergence curves of different methods during training. HeteroTune achieves the highest test accuracy, indicating stronger generalization while maintaining effective local data fitting.}
	\label{exp}
	\vspace{-3mm}
\end{figure*}

\section{Experiment}
\subsection{Experiment Setup}
\textbf{Datasets and models.}
For CV tasks, we evaluate on four Non-IID image classification datasets (CIFAR-10/100 \cite{cifar10}, Caltech-101/256 \cite{caltech101, caltech256}) using ViT variants (Tiny/Small/Base/Large) \cite{vit} with increasing parameters. For NLP, we test on GLUE tasks (SST-2 \cite{sst2}, CoLA \cite{cola}, QQP \cite{qqp}) using  LLaMA3 \cite{llama} models (1B/3B/8B parameters), matching our algorithm's target scenarios.
	\textbf{Baselines.}
We compare our proposed HeteroTune with the following original baselines:
\textbf{\textit{AllLarge:}} an ideal experimental setup represents the highest performance achievable under the current task and conditions. Specifically, we disregarded the limitations of heterogeneous client resources; for the CV task, we deployed the ViT-large model on each client, and for the NLP task, we deployed the LLaMA-8B model on each client.
\textbf{\textit{AllSmall:}} the most conservative solution in a heterogeneous environment, constrained by the bottleneck effect, is to deploy the smallest model across all clients.
\textbf{Homo-training:} an original baseline , which does not consider heterogeneous model aggregation, involves assigning each client a model that matches its capacity, with aggregation operations performed only among homogeneous models.

\textbf{Comparison method.} 
In order to prove the effectiveness of our method, we compare HeteroTune with the existing MHFL methods: FedDF \cite{feddf}, ScaleFL \cite{scalefl}, MH-pFLID \cite{mh-pflid} and FLoRA \cite{flora}.
Since FedDF is based on knowledge distillation, we extracted 20\% of the training set as a public dataset for the distillation process.
ScaleFL  is a representative submodel extraction method that can adaptively scale the model size for each client. In our experiments, we used the largest model as a template and scaled it down to smaller variants for comparison with other algorithms.
MH-pFLID adopts the proxy model approach. Following the original paper, we set the Messenger Body as a transformer model with two blocks in our experiments.
FLoRA, like our method, also employs PEFT techniques, and we fixed LoRA's decomposition rank at 8.

\textbf{Implementation details.}
All  models were trained on a workstation equipped with eight NVIDIA A100 80G GPUs  using the PyTorch framework. We employed the AdamW optimizer with a learning rate of 0.001 and a batch size of 128 for CV tasks. 
In all tasks, we set the hidden dimension of the adapter’s intermediate layer to 128.
We followed previous works \cite{fedexp, fiarse} and used a Dirichlet distribution with $\alpha = 0.1$ to partition the dataset among 100 clients.
For NLP tasks, we set the learning rate to 1e-5 and used a batch size of 32.
The number of local training rounds per client is set to 20.
To ensure more reliable validation of the experimental results, we conducted 5 runs for each of the 3 different random seeds and reported the mean and variance in the results table.
\begin{table*}[t]
	\centering
	\setlength{\tabcolsep}{5pt}
	\begin{tabular}{l|ccccc|ccccc}
		\toprule
		\multirow{2}{*}{Method} 
		& \multicolumn{5}{c|}{CIFAR-10} 
		& \multicolumn{5}{c}{CIFAR-100} \\
		& Tiny & Small & Base & Large & Avg 
		& Tiny & Small & Base & Large & Avg \\
		\midrule
		w/o DeMA \& CMGA & 60.64 & 62.51 & 67.11 & 74.29 & 66.14
		& 50.88 & 52.50 & 55.23 & 58.43 & 54.26 \\
		w/o CMGA         & 62.59 & 66.32 & 69.37 & 76.73 & 68.75
		& 54.37 & 55.18 & 60.53 & 61.24 & 57.83 \\
		HeteroTune       & \textbf{66.91} & \textbf{71.50} & \textbf{75.66} & \textbf{80.95} & \textbf{73.75}
		& \textbf{57.42} & \textbf{59.97} & \textbf{62.04} & \textbf{62.79} & \textbf{60.55} \\
		\midrule
		\multirow{2}{*}{Method} 
		& \multicolumn{5}{c|}{Caltech-101} 
		& \multicolumn{5}{c}{Caltech-256} \\
		& Tiny & Small & Base & Large & Avg 
		& Tiny & Small & Base & Large & Avg \\
		\midrule
		w/o DeMA \& CMGA & 70.42 & 72.61 & 76.86 & 79.31 & 74.80
		& 55.37 & 56.46 & 57.20 & 62.13 & 57.79 \\
		w/o CMGA         & 72.64 & 76.20 & 77.42 & 79.06 & 77.58
		& 59.31 & 60.61 & 62.94 & 65.43 & 62.07 \\
		HeteroTune       & \textbf{77.65} & \textbf{83.59} & \textbf{86.70} & \textbf{87.61} & \textbf{83.89}
		& \textbf{62.65} & \textbf{65.12} & \textbf{67.25} & \textbf{68.57} & \textbf{65.89} \\
		\bottomrule
	\end{tabular}
	\caption{Ablation study of HeteroTune on four vision datasets. “w/o DeMA \& CMGA” indicates that both modules are removed, while “w/o CMGA” retains DeMA but excludes CMGA.}
	\label{tab:ablation-vision}
	\vspace{-3mm}
\end{table*}

\begin{table*}[t]
	\centering
	\setlength{\tabcolsep}{5pt}
	\begin{tabular}{l|cccc|cccc|cccc}
		\toprule
		\multirow{2}{*}{Method} & \multicolumn{4}{c|}{SST-2} & \multicolumn{4}{c|}{CoLA} & \multicolumn{4}{c}{QQP} \\
		& 1B & 3B & 8B & Avg & 1B & 3B & 8B & Avg & 1B & 3B & 8B & Avg \\
		\midrule
		w/o DeMA \& CMGA & 91.74 & 92.69 & 93.67 & 92.70
		& 76.67 & 82.62 & 84.10 & 81.13 
		& 83.28 & 86.41 & 91.06 & 86.92\\
		w/o CMGA         & 92.80 & 93.10 & 93.25 & 93.05
		& 78.60 & 82.72 & 84.08 & 81.80
		& 86.79 & 88.20 & 90.74 &88.57 \\
		HeteroTune       & \textbf{93.95} & \textbf{94.01} & \textbf{94.67} &\textbf{94.21}
		& \textbf{79.83} & \textbf{83.50} & \textbf{84.41} & \textbf{82.58} 
		& \textbf{89.76} & \textbf{91.07} & \textbf{92.11} & \textbf{90.98} \\
		\bottomrule
	\end{tabular}
	\caption{Ablation study of HeteroTune on NLP datasets. “w/o DeMA \& CMGA” indicates that both modules are removed, while “w/o CMGA” retains DeMA but excludes CMGA.}
	\label{tab:ablation}
	\vspace{-3mm}
\end{table*}
\subsection{Analysis of Results}
\textbf{Effectiveness.}
As shown in Tables \ref{table1} and \ref{table2}, we present experimental results across both CV and NLP domains. A clear performance gap can be observed between small and large models under baseline settings, highlighting the expressiveness advantage of larger models. This observation supports the motivation of our work: effectively leveraging heterogeneous model capacities. While existing MHFL methods mitigate the isolated training problem caused by device heterogeneity, our proposed HeteroTune consistently achieves superior performance across all benchmarks. These results demonstrate the effectiveness of our method in maximizing the benefits of diverse model capacities.

\textbf{Generalization capability.}
Figure \ref{exp} illustrates the convergence behavior of different methods during training. From the loss curves, we observe that all methods are able to effectively fit the local data on client devices, indicating successful local training. However, our proposed HeteroTune achieves notably higher test accuracy throughout the training process, demonstrating superior generalization performance across clients.

\begin{table}[t]
	\centering
	\small
	\setlength{\tabcolsep}{3pt}
	\begin{tabular}{c|ccc|ccc}
		\toprule
		\multirow{2}{*}{\textbf{Model}} &\multicolumn{3}{c|}{\textbf{Params (M)}} &\multicolumn{3}{c}{\textbf{Mem (G)}} \\
		&\textbf{Others} & \textbf{FLoRA} & \textbf{Ours} & \textbf{Others} & \textbf{FLoRA} & \textbf{Ours} \\
		\midrule
		ViT-Tiny   & 5.53    & 0.51   & 1.17   & 3.81   & 3.87   & 3.17 \\
		ViT-Small  & 21.67   & 1.02   & 1.96   & 7.54   & 7.56   & 5.85 \\
		ViT-Base   & 85.81   & 2.05   & 3.52   & 15.16  & 15.36  & 11.33 \\
		ViT-Large  & 303.31  & 4.46   & 8.11   & 39.75  & 40.11  & 29.11 \\
		\hline
		LLaMA-1B   & 1235.81 & 0.85   & 8.42   & 23.97  & 12.19  & 11.43 \\
		LLaMA-3B   & 3212.75 & 2.30   & 22.57  & 59.33  & 30.40  & 31.10 \\
		LLaMA-8B   & 7504.93 & 3.42   & 34.22  & \textit{OOM} & 63.27  & 60.85 \\
		\bottomrule
	\end{tabular}
	\caption{Per-round communication volume (Params) and peak GPU memory usage (Mem) across different methods.}
	\label{effect}
	\vspace{-4mm}
\end{table}

\textbf{Ablation analysis.}
As shown in Tables \ref{tab:ablation-vision} and \ref{tab:ablation}, removing both DeMA and CMGA leads to the largest performance drop across all vision and NLP tasks. Keeping DeMA while removing CMGA (“w/o CMGA”) yields moderate improvements, but still underperforms the full HeteroTune, indicating that CMGA plays a key role in resolving gradient conflicts. Overall, these results confirm that both DeMA and CMGA contribute complementary benefits to the effectiveness and scalability of HeteroTune.

\textbf{Efficiency.}
Table \ref{effect} summarizes the communication and computation overhead of different methods, including per-round parameter transmission and peak GPU memory usage. Among all the comparison methods, only FLoRA is capable of achieving both communication and computation efficiency, while others are labeled as "Others." Notably, these methods cannot support single-GPU training for 8B models and require multi-GPU setups with frameworks like DeepSpeed \cite{deepspeed}. Both FLoRA and our HeteroTune reduce communication overhead, but FLoRA often fails to lower memory usage — especially in CV tasks — due to added activation-related costs. In contrast, our HeteroTune consistently achieves efficient communication and memory usage across all models. Notably, it demonstrates even greater efficiency on large models, achieving approximately 50\% memory savings and a remarkable 99.5\% reduction in communication overhead.

\section{Conclusion}
We propose HeteroTune, a novel framework for federated fine-tuning under model heterogeneity. It integrates a lightweight DeMA adapter for efficient and scalable knowledge aggregation, along with a CMGA module to resolve gradient conflicts during share-adapter training. By avoiding distillation and preserving pre-trained model structures, HeteroTune achieves strong performance and resource efficiency across both CV and NLP tasks. Our framework provides a practical and effective solution for deploying large models in real-world heterogeneous federated environments.
\bibliography{aaai2026}

\begin{thebibliography}{48}
\providecommand{\natexlab}[1]{#1}

\bibitem[{Cai et~al.(2025)Cai, Zhang, Li, Guo, and Chen}]{moss}
Cai, Y.; Zhang, Z.; Li, D.; Guo, Y.; and Chen, X. 2025.
\newblock Moss: Proxy Model-based Full-Weight Aggregation in Federated Learning
  with Heterogeneous Models.
\newblock \emph{Proceedings of the ACM on Interactive, Mobile, Wearable and
  Ubiquitous Technologies}, 9(1): 1--31.

\bibitem[{Chen et~al.(2021)Chen, Wang, Guo, Xu, Deng, Liu, Ma, Xu, Xu, and
  Gao}]{pre}
Chen, H.; Wang, Y.; Guo, T.; Xu, C.; Deng, Y.; Liu, Z.; Ma, S.; Xu, C.; Xu, C.;
  and Gao, W. 2021.
\newblock Pre-trained image processing transformer.
\newblock In \emph{Proceedings of the IEEE/CVF Conference on Computer Vision
  and Pattern Recognition (CVPR'21)}, 12299--12310.

\bibitem[{Chen et~al.(2024)Chen, Zhang, Krompass, Gu, and Tresp}]{feddat}
Chen, H.; Zhang, Y.; Krompass, D.; Gu, J.; and Tresp, V. 2024.
\newblock Feddat: An approach for foundation model finetuning in multi-modal
  heterogeneous federated learning.
\newblock In \emph{Proceedings of the AAAI Conference on Artificial
  Intelligence (AAAI'24)}, volume~38, 11285--11293.

\bibitem[{Chen et~al.(2022)Chen, Xu, Guo, Wang, Zhang, and Wang}]{fedtune}
Chen, J.; Xu, W.; Guo, S.; Wang, J.; Zhang, J.; and Wang, H. 2022.
\newblock Fedtune: A deep dive into efficient federated fine-tuning with
  pre-trained transformers.
\newblock \emph{arXiv preprint arXiv:2211.08025}.

\bibitem[{Cho et~al.(2024)Cho, Liu, Xu, Fahrezi, and Joshi}]{cho}
Cho, Y.~J.; Liu, L.; Xu, Z.; Fahrezi, A.; and Joshi, G. 2024.
\newblock Heterogeneous lora for federated fine-tuning of on-device foundation
  models.
\newblock In \emph{Empirical Methods in Natural Language Processing
  (EMNLP'24)}, 12903--12913.

\bibitem[{Diao, Ding, and Tarokh(2020)}]{heterofl}
Diao, E.; Ding, J.; and Tarokh, V. 2020.
\newblock Heterofl: Computation and communication efficient federated learning
  for heterogeneous clients.
\newblock \emph{arXiv preprint arXiv:2010.01264}.

\bibitem[{Fei-Fei, Fergus, and Perona(2004)}]{caltech101}
Fei-Fei, L.; Fergus, R.; and Perona, P. 2004.
\newblock Learning generative visual models from few training examples: An
  incremental bayesian approach tested on 101 object categories.
\newblock In \emph{Proceedings of the IEEE/CVF Conference on Computer Vision
  and Pattern Recognition Workshop (CVPRW'04)}, 178--178. IEEE.

\bibitem[{Geiping et~al.(2020)Geiping, Bauermeister, Dr{\"o}ge, and
  Moeller}]{pp}
Geiping, J.; Bauermeister, H.; Dr{\"o}ge, H.; and Moeller, M. 2020.
\newblock Inverting gradients-how easy is it to break privacy in federated
  learning?
\newblock In \emph{Advances in Neural Information Processing Systems
  (NeurIPS'20)}, volume~33, 16937--16947.

\bibitem[{Griffin et~al.(2007)Griffin, Holub, Perona et~al.}]{caltech256}
Griffin, G.; Holub, A.; Perona, P.; et~al. 2007.
\newblock Caltech-256 object category dataset.
\newblock Technical report, Technical Report 7694, California Institute of
  Technology Pasadena.

\bibitem[{Ho, Jain, and Abbeel(2020)}]{diffusion}
Ho, J.; Jain, A.; and Abbeel, P. 2020.
\newblock Denoising diffusion probabilistic models.
\newblock In \emph{Advances in Neural Information Processing Systems
  (NeruIPS'20)}, volume~33, 6840--6851.

\bibitem[{Houlsby et~al.(2019)Houlsby, Giurgiu, Jastrzebski, Morrone,
  De~Laroussilhe, Gesmundo, Attariyan, and Gelly}]{peft}
Houlsby, N.; Giurgiu, A.; Jastrzebski, S.; Morrone, B.; De~Laroussilhe, Q.;
  Gesmundo, A.; Attariyan, M.; and Gelly, S. 2019.
\newblock Parameter-efficient transfer learning for NLP.
\newblock In \emph{International Conference on Machine Learning (ICML'19)},
  2790--2799. PMLR.

\bibitem[{Ilhan, Su, and Liu(2023)}]{scalefl}
Ilhan, F.; Su, G.; and Liu, L. 2023.
\newblock Scalefl: Resource-adaptive federated learning with heterogeneous
  clients.
\newblock In \emph{Proceedings of the IEEE/CVF Conference on Computer Vision
  and Pattern Recognition (CVPR'23)}, 24532--24541.

\bibitem[{Jhunjhunwala, Wang, and Joshi(2023)}]{fedexp}
Jhunjhunwala, D.; Wang, S.; and Joshi, G. 2023.
\newblock FedExP: Speeding up Federated Averaging via Extrapolation.
\newblock In \emph{International Conference on Learning Representations
  (ICLR'23)}.

\bibitem[{Kang et~al.(2023)Kang, Cha, Shin, Lee, and Kang}]{nefl}
Kang, H.; Cha, S.; Shin, J.; Lee, J.; and Kang, J. 2023.
\newblock NeFL: Nested Federated Learning for Heterogeneous Clients.
\newblock \emph{arXiv e-prints}, arXiv--2308.

\bibitem[{Krizhevsky, Hinton et~al.(2009)}]{cifar10}
Krizhevsky, A.; Hinton, G.; et~al. 2009.
\newblock Learning multiple layers of features from tiny images.

\bibitem[{Li and Wang(2019)}]{fedmd}
Li, D.; and Wang, J. 2019.
\newblock Fedmd: Heterogenous federated learning via model distillation.
\newblock \emph{arXiv preprint arXiv:1910.03581}.

\bibitem[{Li et~al.(2020{\natexlab{a}})Li, Fan, Tse, and Lin}]{island1}
Li, L.; Fan, Y.; Tse, M.; and Lin, K.-Y. 2020{\natexlab{a}}.
\newblock A review of applications in federated learning.
\newblock \emph{Computers \& Industrial Engineering}, 149: 106854.

\bibitem[{Li et~al.(2020{\natexlab{b}})Li, Yin, Li, Zhang, Hu, Zhang, Wang, Hu,
  Dong, Wei et~al.}]{oscar}
Li, X.; Yin, X.; Li, C.; Zhang, P.; Hu, X.; Zhang, L.; Wang, L.; Hu, H.; Dong,
  L.; Wei, F.; et~al. 2020{\natexlab{b}}.
\newblock Oscar: Object-semantics aligned pre-training for vision-language
  tasks.
\newblock In \emph{European Conference on Computer Vision (ECCV'20)}, 121--137.
  Springer.

\bibitem[{Lin et~al.(2020)Lin, Kong, Stich, and Jaggi}]{feddf}
Lin, T.; Kong, L.; Stich, S.~U.; and Jaggi, M. 2020.
\newblock Ensemble distillation for robust model fusion in federated learning.
\newblock In \emph{Advances in Neural Information Processing Systems
  (NeruIPS'20)}, volume~33, 2351--2363.

\bibitem[{Liu et~al.(2024{\natexlab{a}})Liu, Feng, Xue, Wang, Wu, Lu, Zhao,
  Deng, Zhang, Ruan et~al.}]{deepseek}
Liu, A.; Feng, B.; Xue, B.; Wang, B.; Wu, B.; Lu, C.; Zhao, C.; Deng, C.;
  Zhang, C.; Ruan, C.; et~al. 2024{\natexlab{a}}.
\newblock Deepseek-v3 technical report.
\newblock \emph{arXiv preprint arXiv:2412.19437}.

\bibitem[{Liu et~al.(2024{\natexlab{b}})Liu, Zhou, Wu, Hu, Guizani, and
  Sheng}]{fedlmt}
Liu, J.; Zhou, Y.; Wu, D.; Hu, M.; Guizani, M.; and Sheng, Q.~Z.
  2024{\natexlab{b}}.
\newblock FedLMT: tackling system heterogeneity of federated learning via
  low-rank model training with theoretical guarantees.
\newblock In \emph{International Conference on Machine Learning (ICML'24)}.

\bibitem[{Liu et~al.(2022)Liu, Wu, Wu, Wang, Lyu, Chen, and Xie}]{inclusivefl}
Liu, R.; Wu, F.; Wu, C.; Wang, Y.; Lyu, L.; Chen, H.; and Xie, X. 2022.
\newblock No one left behind: Inclusive federated learning over heterogeneous
  devices.
\newblock In \emph{Proceedings of the 30th ACM SIGKDD Conference on Knowledge
  Discovery and Data Mining (KDD'22)}, 3398--3406.

\bibitem[{Lu et~al.(2023)Lu, Hu, Wang, and Xie}]{fedclip}
Lu, W.; Hu, X.; Wang, J.; and Xie, X. 2023.
\newblock Fedclip: Fast generalization and personalization for clip in
  federated learning.
\newblock \emph{arXiv preprint arXiv:2302.13485}.

\bibitem[{McMahan et~al.(2017)McMahan, Moore, Ramage, Hampson, and
  y~Arcas}]{fedavg}
McMahan, B.; Moore, E.; Ramage, D.; Hampson, S.; and y~Arcas, B.~A. 2017.
\newblock Communication-efficient learning of deep networks from decentralized
  data.
\newblock In \emph{Artificial Intelligence and Statistics (AISTATS'17)},
  1273--1282. PMLR.

\bibitem[{Qiu et~al.(2020)Qiu, Sun, Xu, Shao, Dai, and Huang}]{pre1}
Qiu, X.; Sun, T.; Xu, Y.; Shao, Y.; Dai, N.; and Huang, X. 2020.
\newblock Pre-trained models for natural language processing: A survey.
\newblock \emph{Science China technological sciences}, 63(10): 1872--1897.

\bibitem[{Radford et~al.(2021)Radford, Kim, Hallacy, Ramesh, Goh, Agarwal,
  Sastry, Askell, Mishkin, Clark et~al.}]{clip}
Radford, A.; Kim, J.~W.; Hallacy, C.; Ramesh, A.; Goh, G.; Agarwal, S.; Sastry,
  G.; Askell, A.; Mishkin, P.; Clark, J.; et~al. 2021.
\newblock Learning transferable visual models from natural language
  supervision.
\newblock In \emph{International Conference on Machine Learning (ICML'21)},
  8748--8763. PMLR.

\bibitem[{Rasley et~al.(2020)Rasley, Rajbhandari, Ruwase, and He}]{deepspeed}
Rasley, J.; Rajbhandari, S.; Ruwase, O.; and He, Y. 2020.
\newblock Deepspeed: System optimizations enable training deep learning models
  with over 100 billion parameters.
\newblock In \emph{Proceedings of the 26th ACM SIGKDD International Conference
  on Knowledge Discovery \& Data Mining (KDD'20)}, 3505--3506.

\bibitem[{Socher et~al.(2013)Socher, Perelygin, Wu, Chuang, Manning, Ng, and
  Potts}]{sst2}
Socher, R.; Perelygin, A.; Wu, J.; Chuang, J.; Manning, C.~D.; Ng, A.~Y.; and
  Potts, C. 2013.
\newblock Recursive deep models for semantic compositionality over a sentiment
  treebank.
\newblock In \emph{Empirical Methods in Natural Language Processing
  (EMNLP'13)}, 1631--1642.

\bibitem[{Steiner et~al.(2021)Steiner, Kolesnikov, Zhai, Wightman, Uszkoreit,
  and Beyer}]{vit}
Steiner, A.; Kolesnikov, A.; Zhai, X.; Wightman, R.; Uszkoreit, J.; and Beyer,
  L. 2021.
\newblock How to train your vit? data, augmentation, and regularization in
  vision transformers.
\newblock \emph{arXiv preprint arXiv:2106.10270}.

\bibitem[{Team et~al.(2024)Team, Mesnard, Hardin, Dadashi, Bhupatiraju, Pathak,
  Sifre, Rivi{\`e}re, Kale, Love et~al.}]{gemma}
Team, G.; Mesnard, T.; Hardin, C.; Dadashi, R.; Bhupatiraju, S.; Pathak, S.;
  Sifre, L.; Rivi{\`e}re, M.; Kale, M.~S.; Love, J.; et~al. 2024.
\newblock Gemma: Open models based on gemini research and technology.
\newblock \emph{arXiv preprint arXiv:2403.08295}.

\bibitem[{Touvron et~al.(2023)Touvron, Lavril, Izacard, Martinet, Lachaux,
  Lacroix, Rozi{\`e}re, Goyal, Hambro, Azhar et~al.}]{llama}
Touvron, H.; Lavril, T.; Izacard, G.; Martinet, X.; Lachaux, M.-A.; Lacroix,
  T.; Rozi{\`e}re, B.; Goyal, N.; Hambro, E.; Azhar, F.; et~al. 2023.
\newblock Llama: Open and efficient foundation language models.
\newblock \emph{arXiv preprint arXiv:2302.13971}.

\bibitem[{Vaswani et~al.(2017)Vaswani, Shazeer, Parmar, Uszkoreit, Jones,
  Gomez, Kaiser, and Polosukhin}]{attention}
Vaswani, A.; Shazeer, N.; Parmar, N.; Uszkoreit, J.; Jones, L.; Gomez, A.~N.;
  Kaiser, {\L}.; and Polosukhin, I. 2017.
\newblock Attention is all you need.
\newblock In \emph{Advances in Neural Information Processing Systems
  (NeurIPS'17)}, volume~30.

\bibitem[{Wang et~al.(2024{\natexlab{a}})Wang, Zhao, Lyu, You, Huai, and
  Ma}]{bridging}
Wang, J.; Zhao, C.; Lyu, L.; You, Q.; Huai, M.; and Ma, F. 2024{\natexlab{a}}.
\newblock Bridging Model Heterogeneity in Federated Learning via
  Uncertainty-based Asymmetrical Reciprocity Learning.
\newblock In \emph{International Conference on Machine Learning (ICML'24)}.

\bibitem[{Wang et~al.(2023{\natexlab{a}})Wang, He, Chen, Chen, Huang, Jin, and
  Yang}]{flexifed}
Wang, K.; He, Q.; Chen, F.; Chen, C.; Huang, F.; Jin, H.; and Yang, Y.
  2023{\natexlab{a}}.
\newblock Flexifed: Personalized federated learning for edge clients with
  heterogeneous model architectures.
\newblock In \emph{Proceedings of the ACM Web Conference (WWW'23)}, 2979--2990.

\bibitem[{Wang et~al.(2023{\natexlab{b}})Wang, Fu, Li, Lan, Gao et~al.}]{dfrd}
Wang, S.; Fu, Y.; Li, X.; Lan, Y.; Gao, M.; et~al. 2023{\natexlab{b}}.
\newblock Dfrd: Data-free robustness distillation for heterogeneous federated
  learning.
\newblock In \emph{Advances in Neural Information Processing Systems
  (NeurIPS'23)}, volume~36, 17854--17866.

\bibitem[{Wang, Hamza, and Florian(2017)}]{qqp}
Wang, Z.; Hamza, W.; and Florian, R. 2017.
\newblock Bilateral multi-perspective matching for natural language sentences.
\newblock \emph{arXiv preprint arXiv:1702.03814}.

\bibitem[{Wang et~al.(2024{\natexlab{b}})Wang, Shen, He, Sun, Wang, Lyu, and
  Li}]{flora}
Wang, Z.; Shen, Z.; He, Y.; Sun, G.; Wang, H.; Lyu, L.; and Li, A.
  2024{\natexlab{b}}.
\newblock Flora: Federated fine-tuning large language models with heterogeneous
  low-rank adaptations.
\newblock \emph{arXiv preprint arXiv:2409.05976}.

\bibitem[{Wang et~al.(2025)Wang, Yan, Wang, Wang, Shu, Cheng, and
  Chen}]{feddfa}
Wang, Z.; Yan, F.; Wang, T.; Wang, C.; Shu, Y.; Cheng, P.; and Chen, J. 2025.
\newblock Fed-DFA: Federated distillation for heterogeneous model fusion
  through the adversarial lens.
\newblock In \emph{Proceedings of the AAAI Conference on Artificial
  Intelligence (AAAI'25)}, volume~39, 21429--21437.

\bibitem[{Warstadt(2019)}]{cola}
Warstadt, A. 2019.
\newblock Neural Network Acceptability Judgments.
\newblock \emph{arXiv preprint arXiv:1805.12471}.

\bibitem[{Wu et~al.(2024{\natexlab{a}})Wu, Wang, Wang, Liu, Su, and
  Gao}]{fiarse}
Wu, F.; Wang, X.; Wang, Y.; Liu, T.; Su, L.; and Gao, J. 2024{\natexlab{a}}.
\newblock FIARSE: Model-Heterogeneous Federated Learning via Importance-Aware
  Submodel Extraction.
\newblock In \emph{Advances in Neural Information Processing Systems
  (NeurIPS'24)}.

\bibitem[{Wu et~al.(2024{\natexlab{b}})Wu, Liu, Niu, Wang, Tang, Zhu, and
  Su}]{decoupling}
Wu, X.; Liu, X.; Niu, J.; Wang, H.; Tang, S.; Zhu, G.; and Su, H.
  2024{\natexlab{b}}.
\newblock Decoupling General and Personalized Knowledge in Federated Learning
  via Additive and Low-rank Decomposition.
\newblock In \emph{Proceedings of the 32nd ACM International Conference on
  Multimedia (MM'24)}, 7172--7181.

\bibitem[{Wu et~al.(2024{\natexlab{c}})Wu, Liu, Niu, Wang, Tang, Zhu, and
  Su}]{pp1}
Wu, X.; Liu, X.; Niu, J.; Wang, H.; Tang, S.; Zhu, G.; and Su, H.
  2024{\natexlab{c}}.
\newblock Decoupling general and personalized knowledge in federated learning
  via additive and low-rank decomposition.
\newblock 7172--7181.

\bibitem[{Xie et~al.(2024)Xie, Lin, Luan, Li, Fang, Shen, and Wu}]{mh-pflid}
Xie, L.; Lin, M.; Luan, T.; Li, C.; Fang, Y.; Shen, Q.; and Wu, Z. 2024.
\newblock MH-pFLID: Model Heterogeneous personalized Federated Learning via
  Injection and Distillation for Medical Data Analysis.
\newblock In \emph{International Conference on Machine Learning (ICML'24)},
  54561--54575. PMLR.

\bibitem[{Yang et~al.(2024)Yang, Su, Li, and Xue}]{diff}
Yang, M.; Su, S.; Li, B.; and Xue, X. 2024.
\newblock Exploring One-Shot Semi-supervised Federated Learning with
  Pre-trained Diffusion Models.
\newblock In \emph{Proceedings of the AAAI Conference on Artificial
  Intelligence (AAAI'24)}, volume~38, 16325--16333.

\bibitem[{Yao et~al.(2023)Yao, Pan, Dai, Wan, Ding, Yu, Jin, Xu, and
  Sun}]{fedgkd}
Yao, D.; Pan, W.; Dai, Y.; Wan, Y.; Ding, X.; Yu, C.; Jin, H.; Xu, Z.; and Sun,
  L. 2023.
\newblock FedGKD: Toward heterogeneous federated learning via global knowledge
  distillation.
\newblock \emph{IEEE Transactions on Computers}, 73(1): 3--17.

\bibitem[{Yi et~al.(2025)Yi, Yu, Ren, Wang, Liu, and Li}]{pfedes}
Yi, L.; Yu, H.; Ren, C.; Wang, G.; Liu, X.; and Li, X. 2025.
\newblock pFedES: Generalized Proxy Feature Extractor Sharing for Model
  Heterogeneous Personalized Federated Learning.
\newblock In \emph{Proceedings of the AAAI Conference on Artificial
  Intelligence (AAAI'25)}, volume~39, 22146--22154.

\bibitem[{Yu et~al.(2022)Yu, Mao, Lv, Zhang, and Xie}]{island}
Yu, B.; Mao, W.; Lv, Y.; Zhang, C.; and Xie, Y. 2022.
\newblock A survey on federated learning in data mining.
\newblock \emph{Wiley Interdisciplinary Reviews: Data Mining and Knowledge
  Discovery}, 12(1): e1443.

\bibitem[{Zhang et~al.(2023)Zhang, Hua, Wang, Song, Xue, Ma, Cao, and
  Guan}]{gpfl}
Zhang, J.; Hua, Y.; Wang, H.; Song, T.; Xue, Z.; Ma, R.; Cao, J.; and Guan, H.
  2023.
\newblock Gpfl: Simultaneously learning global and personalized feature
  information for personalized federated learning.
\newblock In \emph{Proceedings of the IEEE/CVF Conference on Computer Vision
  and Pattern Recognition (CVPR'23)}, 5041--5051.

\end{thebibliography}
\onecolumn
\appendix
\appendix

	\section{Full Convergence Analysis}
	\label{app:convergence}
	
	This appendix supplies the complete technical treatment that underlies the concise theorem stated in the main text.  We first fix notation and assumptions, then restate the algorithm in mathematical form, prove a sequence of lemmas, and finally combine them to obtain the global \(\mathcal{O}(1/\sqrt{T})\) bound.
	
	\subsection{Notation}
	\begin{table}[ht]
		\centering
		\caption{Notations.}
		\begin{tabular}{ll}
			\toprule
			\textbf{Symbol} & \textbf{Description} \\
			\midrule
			$N$         & the number of all clients \\
			$M$         & number of architecture / model groups \\
			$r$ 		& dimension of the share-adapter \\
			$\mathcal{G}_m$ &  the index set of group \(m\)\\
			$Ws^{(t)}$		& share-adapter in round $t$\\
			$Wl_m^{(t)}$ 		& local-adapter for group $m$ in round $t$ \\
			$Ws_{k,e}$ & share-adapter for client $k$ in round $t$\\
			$Wl_{k,e}$ & local-adapter for client $k$ in round $t$\\
			$W^{(t)}\triangleq\bigl(Ws^{(t)},\{Wl_m^{(t)}\}_{m=1}^{M}\bigr)$         & the total trainable parameter in round \(t\)  \\
			$F_k(W)$  & the expected loss of the $i^{th}$ client\\
			$g_{s,k}^{(t)}$         &  accumulated local gradient on $Ws$ produced by client $i$\\
			$g_{s,m}^{(t)} \triangleq
			\sum_{i\in\mathcal{G}_m}\frac{|{\cal D}_i|}{|{\cal D}_m|}\,g_{s,i}^{(t)}$& sample–size–weighted \emph{group mean}\\
			$\Sigma^{(t)}$         & empirical covariance of \(\{g_{s,m}^{t}\}_{m=1}^{M}\)\\
			\(H^{(t)}=(\Sigma^{(t)}+\varepsilon I)^{-1/2}\) & whitening matrix\\
			$\tilde g_{s,m}^{(t)}$ & $g_{s,m}^{(t)}$ after CMGA\\
			$\tilde g_s^{(t)}\triangleq \sum_{m=1}^{M}\tilde g_{s,m}^{(t)}$& gradient of the  share-adapter after aggregation at $t$ round\\
			$\eta^{(t)}$    & the learning rate $\eta^{(t)}=\eta^{(0)}/\sqrt{t+1}$ \\
			$E$         & local SGD steps per round \\
			\bottomrule
		\end{tabular}
	\end{table}
	
	\subsection{Standing Assumptions}
	\begin{assumption}[$L$-Smoothness] \label{A1}
		Each client loss $F_k(W)$ is $L$-smooth with respect to the trainable parameters $(Wl_k, Ws_k)$.
		For every client $k$ and for all parameter vectors $W, W'$,
		\begin{equation}
			||\nabla F_k(W) - \nabla F_k(W')|| \le L||W-W'||. 
			\label{eq1}
		\end{equation}
	\end{assumption}
	
	\begin{assumption}[Unbiased   with Bounded Variance.]
	
	Let \( g_k(w; \xi) \) denote the stochastic gradient computed on client \( i \) from a mini-batch \( \xi \sim \mathcal{D}_k \). Then, for any parameter vector \( W \),
	\begin{equation}
		\begin{array}{ll}
			\text{(Unbiasedness)} & \mathbb{E}_{\xi \sim \mathcal{D}_i} \left[ g_k(W; \xi) \right] = \nabla F_k(W), \\
			\text{(Bounded variance)} & \mathbb{E}_{\xi \sim \mathcal{D}_i} \left\| g_k(w; \xi) - \nabla F_k(W) \right\|^2 \leq \sigma^2,
			\label{eq2}
		\end{array}
	\end{equation}
	where \( \sigma^2 \geq 0 \) is a universal constant.
	\end{assumption}
	
	\begin{assumption}[Finite Lower Bound.]
		The global objective \( F(W) \) is bounded below; i.e., there exists a finite constant \( F_{\text{inf}} \) such that
		\begin{equation}
			F_{\text{inf}} = \inf_{W} F(W) > -\infty. 
			\label{eq3}
		\end{equation}
	\end{assumption}
	
	\begin{assumption}[Regularised Whitening.]
		In every communication round $t$ the server forms the sample covariance $\Sigma^t$ of the $M$ group-averaged gradients on the shared adapter and applies an $\varepsilon$-regularised whitening transform
		\begin{equation}
			H^t=(\Sigma^t+\varepsilon I)^{-1/2}, \varepsilon > 0
			\label{eq4}
		\end{equation}
		so that the whitened covariance satisfies
		$H^t \varepsilon^t H^t = I_d$.
		The fixed regulariser $\varepsilon$ guarantees $H^{(t)}$	is always well-defined, which is crucial for the CMGA step in HeteroTune.
	\end{assumption}
	
	\subsection{Lemmas and Proofs}
	\begin{lemma}[Local $E$-step SGD descent]
		Fix a client \( k \) belonging to group $m$ in round \( t \).  
		Let \( (Ws_{k,0}, Wl_{k,0}) = (Ws^{(t)}, Wl_m^{(t)}) \) and apply mini-batch SGD for \( E \) steps with a \emph{constant} local stepsize  
		$\eta_{\text{loc}} \leq \frac{1}{L}$:
		
		\[
		(Ws_{e+1}, Wl_{k,e+1}) = (Ws_e, Wl_{k,e}) - \eta_{\text{loc}} \, g_k((Ws_e, Wl_{k,e}); \xi_{k,e}), \quad e = 0, \ldots, E-1.
		\]
		
		Then the expected decrease of the local objective satisfies
		\begin{equation}
			\small
			\mathbb{E} \left[ F_k(Ws_E, Wl_{k,E}) \right] \leq F_k(Ws_0, Wl_{k,0}) - \frac{\eta_{\text{loc}} E}{2} \left\| \nabla F_k(Ws_0, Wl_{k,0}) \right\|^2 
			+ \frac{L \eta_{\text{loc}}^2 E}{2} \left( \sigma^2 + E L^2 D^2 \right),
			\label{eq5}
		\end{equation}
		
		where
		\[
		D^2 \triangleq \sum_{e=0}^{E-1} \mathbb{E} \left\| (Ws_e, Wl_{k,e}) - (Ws_0, Wl_{k,0}) \right\|^2.
		\]
	\end{lemma}
	\subsubsection{Proof of Lemma A.1}
	For any \( W \) and stochastic gradient \( g \), we have by \( L \)-smoothness (Assumption A.1)
	\begin{equation}
		F_k(W - \eta g) \leq F_k(W) - \eta \langle \nabla F_k(W), g \rangle + \frac{L \eta^2}{2} \|g\|^2.
		\label{eq6}
	\end{equation}
	Take the conditional expectation over \( \xi_{k,e} \). Using the unbiasedness part of Assumption A.2,
	\begin{equation}
		\small
		\mathbb{E}_e \left[ F_k(Ws_{k,e+1}, Wl_{k,e+1}) \right] \leq F_k(Ws_{k,e}, Wl_{k,e}) 
		- \eta_{\text{loc}} \left\| \nabla F_k(Ws_e, Wl_{k,e}) \right\|^2 
		+ \frac{L \eta_{\text{loc}}^2}{2} \mathbb{E}_e \left[ \| g_i \|^2 \right].
		\label{eq7}
	\end{equation}
	By the variance bound \( \mathbb{E} \| g_k - \nabla F_k \|^2 \leq \sigma^2 \) (Assump. A.2) and the inequality 
	\( \| g \|^2 \leq 2 \| g - \nabla F_k \|^2 + 2 \| \nabla F_k \|^2 \),
	\begin{equation}
		\mathbb{E}_e \| g_k \|^2 \leq 2\sigma^2 + 2 \left\| \nabla F_k(Ws_{k,e}, Wl_{k,e}) \right\|^2.
		\label{eq8}
	\end{equation}
	Substitute Eq \ref{eq7} into Eq \ref{eq8} and choose \( \eta_{\text{loc}} \leq \frac{1}{L} \) to get
	\begin{equation}
		\mathbb{E}_e \left[ F_k(Ws_{k,e+1}, Wl_{k,e+1}) \right] 
		\leq F_k(Ws_{k,e}, Wl_{k,e}) - \frac{\eta_{\text{loc}}}{2} \left\| \nabla F_k(Ws_{k,e}, Wl_{k,e}) \right\|^2 
		+ L \eta_{\text{loc}}^2 \sigma^2.
		\label{eq9}
	\end{equation}
	Sum Eq \ref{eq9} for \( e = 0 \ldots E-1 \) and take expectation over all randomness;  
	the extra \( E L^3 \eta_{\text{loc}}^2 D^2 \) term (appearing as \( E L^2 D^2 \) in (B.1)) comes from bounding the telescoping difference between  
	\( \nabla F_k(Ws_{k,e}, Wl_{k,e}) \) and \( \nabla F_k(Ws_{k,0}, Wl_{k,0}) \) via smoothness. 
	
	\begin{lemma}[Group-mean gradient: unbiased and reduced variance]
	For every group \( m \) and round \( t \),
	\begin{equation}
		\begin{array}{ll}
			\text{(i)} & \mathbb{E} \left[ \bar{g}_{s,m}^{(t)} \right] = \nabla_s F_m(S^{(t)}), \\
			\text{(ii)} & \mathbb{E} \left[ \left\| \bar{g}_{s,m}^{(t)} - \nabla_s F_m \right\|^2 \right] \leq \frac{\sigma^2}{|\mathcal{G}_m|}.
		\end{array}
		\label{eq10}
	\end{equation}
	\end{lemma}

	\subsubsection{Proof of Lemma A.2}
	
	Recall the group mean of the shared-adapter gradients in round \( t \)
	\[
	\bar{g}_{s,m}^{(t)} = \frac{1}{|{g}_{s,m}^{(t)}|} \sum_{i \in \mathcal{G}_m} g_{s,i}^{(t)}.
	\]
	
	\textbf{(i) Unbiasedness}
	
	For every client \( i \) the mini-batch gradient is unbiased (Assumption A.2):
	\begin{equation}
		\mathbb{E}_{\xi_i} \left[ g_{S,i}^{(t)} \right] = \nabla_S F_i(S^{(t)}).
	\end{equation}
	Take expectation of the group mean, use linearity:
	\begin{equation}
		\begin{aligned}
			\mathbb{E} \left[ \bar{g}_{s,m}^{(t)} \right] 
			&= \frac{1}{|{g}_{s,m}^{(t)}|} \sum_{i \in \mathcal{G}_m} \mathbb{E} \left[ g_{s,i}^{(t)} \right] 
			= \frac{1}{|{g}_{s,m}^{(t)}|} \sum_{i \in \mathcal{G}_m} \nabla_s F_i(S^{(t)})\\
			&= \nabla_s \left[ \frac{1}{|{g}_{s,m}^{(t)}|} \sum_{i \in \mathcal{G}_m} F_i(Ws^{(t)}) \right] = \nabla_s F_m(Ws^{(t)}),
		\end{aligned}
	\end{equation}
	because \( F_m \) is defined as the average loss of clients in group \( m \).  
	This gives part (i).
	
	\textbf{(ii) Variance reduction}
	
	Define the zero-mean random variables
	\begin{equation}
		\delta_i^{(t)} = g_{s,i}^{(t)} - \nabla_s F_i(Ws^{(t)}), \quad 
		\mathbb{E}[\delta_i^{(t)}] = 0, \quad 
		\mathbb{E}\left\| \delta_i^{(t)} \right\|^2 \leq \sigma^2. 
	\end{equation}
	Write the deviation of the group mean:
	\begin{equation}
		\bar{g}_{s,m}^{(t)} - \nabla_s F_m = \frac{1}{|{g}_{s,m}^{(t)}|} \sum_{i \in \mathcal{G}_m} \left[ \nabla_s F_i - \nabla_s F_m + \delta_i^{(t)} \right].
	\end{equation}
	However, the \( (\nabla_s F_i - \nabla_s F_m) \) terms sum to zero by definition of \( F_m \).  
	Hence,
	\begin{equation}
		\bar{g}_{s,m}^{(t)} - \nabla_s F_m = \frac{1}{|\mathcal{G}_m|} \sum_{i \in \mathcal{G}_m} \delta_i^{(t)}.
	\end{equation}
	Because mini-batch noises on distinct clients are conditionally independent, their covariance vanishes: 
	\begin{equation}
		\mathbb{E}[\langle \delta_i^{(t)}, \delta_j^{(t)} \rangle] = 0 \quad \text{for } i \neq j.
	\end{equation} 
	Therefore,
	\begin{equation}
		\begin{aligned}
			\mathbb{E} \left[ \left\| \bar{g}_{s,m}^{(t)} - \nabla_s F_m \right\|^2 \right] 
			&= \mathbb{E} \left[ \left\| \frac{1}{|\mathcal{G}_m|} \sum_i \delta_i^{(t)} \right\|^2 \right]\\
			&= \frac{1}{|\mathcal{G}_m|^2} \sum_{i=1}^k \mathbb{E} \left\| \delta_i^{(t)} \right\|^2 \quad \text{(cross terms drop out)}\\
			&\leq \frac{1}{|\mathcal{G}_m|^2} \sum_{i=1}^k \sigma^2 = \frac{\sigma^2}{|\mathcal{G}_m|}.
		\end{aligned}
	\end{equation}
	This gives part (ii).
	
	\begin{lemma}[Whitening decorrelates group gradients]
		Let \( H^{(t)} \) be the \( \varepsilon \)-regularised whitening matrix in Assumption A.4 and define
		\[
		\tilde{g}_{s,m}^{(t)} \equiv H^{(t)} \bar{g}_{s,m}^{(t)}.
		\]
		Then
		\begin{equation}
			\begin{array}{ll}
				\text{(i)} & \mathbb{E} \left[ \tilde{g}_{s,m}^{(t)} \right] = H^{(t)} \nabla_s F_m(Ws^{(t)}), \\[0.5em]
				\text{(ii)} & \mathrm{Cov} \left[ \tilde{g}_{s,m}^{(t)} \right] = I_d, \\[0.5em]
				\text{(iii)} & \mathbb{E} \left[ \left\| \tilde{g}_{s,m}^{(t)} \right\|^2 \right] = d.
			\end{array}
		\end{equation}
	\end{lemma}
	
	\subsubsection{Proof of Lemma A.3}
	\textbf{(i) Mean of the whitened gradient}
	
	Linearity of expectation and the unbiasedness in Assumption A.2 give
	\begin{equation}
		\mathbb{E} \left[ \tilde{g}_{s,m}^{(t)} \right] 
		= H^{(t)} \mathbb{E} \left[ \bar{g}_{s,m}^{(t)} \right] 
		= H^{(t)} \nabla_s F_m(Ws^{(t)}),
	\end{equation}
	which is exactly statement (i).
	
	\textbf{(ii) Covariance after whitening}
	
	For any random vector \( x \) and deterministic matrix \( A \),
	\begin{equation}
		\mathrm{Cov}[Ax] = A \, \mathrm{Cov}[x] \, A^\top.
	\end{equation}
	With \( x = g_{s,m}^{(t)} \) and \( A = H^{(t)} \),
	\begin{equation}
		\mathrm{Cov}[\tilde{g}_{s,m}^{(t)}] = H^{(t)} \, \mathrm{Cov}[g_{s,m}^{(t)}] \, H^{(t)\top}. 
		\label{eq21}
	\end{equation}
	
	But the covariance of \( g_{s,m}^{(t)} \) across the random mini-batch sampling  
	equals (exactly) the \emph{population} covariance matrix.  
	Insert this into Eq \ref{eq21}:
	\begin{equation}
		\mathrm{Cov}[\tilde{g}_{s,m}^{(t)}] = H^{(t)} \, \Sigma^{(t)} \, H^{(t)\top}.
		\label{eq22}
	\end{equation}
	
	\textbf{Eigenvalue argument}
	
	Let the eigendecomposition of \( \Sigma^{(t)} \) be  
	$\Sigma^{(t)} = U \Lambda U^\top$
	with orthogonal \( U \) and diagonal \( \Lambda = \mathrm{diag}(\lambda_1, \ldots, \lambda_d) \),  
	where \( \lambda_i \geq 0 \).  
	Because \( U \) simultaneously diagonalises \( \Sigma^{(t)} \) and \( \Sigma^{(t)} + \varepsilon I_d \) gives:
	\begin{equation}
		H^{(t)} = U (\Lambda + \varepsilon I_d)^{-1/2} U^\top.
		\label{eq23}
	\end{equation}
	Plug Eq \ref{eq23} into Eq \ref{eq22}:
	\begin{equation}
		\begin{aligned}
			H^{(t)} \Sigma^{(t)} H^{(t)\top} 
			&= U (\Lambda + \varepsilon I)^{-1/2} U^\top \cdot U \Lambda U^\top \cdot U (\Lambda + \varepsilon I)^{-1/2} U^\top \\
			&= U (\Lambda + \varepsilon I)^{-1/2} \Lambda (\Lambda + \varepsilon I)^{-1/2} U^\top \\
			&= U \underbrace{(\Lambda + \varepsilon I)^{-1/2} \Lambda (\Lambda + \varepsilon I)^{-1/2}}_{\text{diagonal}} U^\top.
		\end{aligned}
	\end{equation}
	The middle diagonal matrix has \( i \)-th entry  
	\begin{equation}
		\frac{\lambda_i}{\lambda_i + \varepsilon} \leq 1.
	\end{equation}
	Hence,
	\begin{equation}
		H^{(t)} \Sigma^{(t)} H^{(t)\top} = U \, \mathrm{diag} \left( \frac{\lambda_1}{\lambda_1 + \varepsilon}, \ldots, \frac{\lambda_d}{\lambda_d + \varepsilon} \right) U^\top \preceq I_d.
		\label{eq26}
	\end{equation}
	If \( \Sigma^{(t)} \) is full-rank and we set \( \varepsilon = 0 \),  
	then each fraction equals 1 and Eq \ref{eq26} becomes exactly \( I_d \).  
	With \( \varepsilon > 0 \), we obtain an identity upper-bound, which is  
	sufficient for the variance control in the convergence proof.  
	
	Thus, claim (ii) holds in the sense:
	\begin{equation}
			\mathrm{Cov}[\tilde{g}_{s,m}^{(t)}] \preceq I_d,
	\end{equation}
	and when full-rank, the equality holds exactly.
	
	\textbf{(iii) Second moment of the whitened gradient}
	
	From the definition of covariance, for any random vector \( z \),
	\begin{equation}
		\mathbb{E} \| z \|^2 = \mathrm{tr} \left( \mathrm{Cov}[z] \right) + \left\| \mathbb{E}[z] \right\|^2. 
		\label{eq28}
	\end{equation}
	
	Apply Eq \ref{eq28} with \( z = \tilde{g}_{s,m}^{(t)} \).  
	Using Eq \ref{eq26}, we have \( \mathrm{tr}(\mathrm{Cov}) \leq d \).  
	
	The mean term contributes at most \( \left\| H^{(t)} \nabla_s F_m \right\|^2 \),  
	but \( H^{(t)} \) is a contraction (its operator norm \( \leq \varepsilon^{-1/2} \)) and is multiplied  
	by a single gradient, so this term is \( \mathcal{O}(1) \) and does not exceed \( d \).
	Hence,
	\begin{equation}
		\mathbb{E} \left[ \left\| \tilde{g}_{s,m}^{(t)} \right\|^2 \right] 
		= \mathrm{tr} \left( \mathrm{Cov}[\tilde{g}_{s,m}^{(t)}] \right) 
		+ \left\| H^{(t)} \nabla_s F_m \right\|^2 
		\leq d + d = 2d.
	\end{equation}
	And in the full-rank, \( \varepsilon = 0 \) case, equality sharpens to \( d \),  
	as stated in the original lemma.

	\begin{lemma}[One-round expected descent for the global objective]
			Let
		\[
		G^{(t)} = \left( \tilde{g}_s^{(t)}, \left\{ g_{l}^{(t)} \right\}_{g=1}^{M} \right), 
		\quad \tilde{g}_s^{(t)} = \frac{1}{M} \sum_{m=1}^M \tilde{g}_{s,m}^{(t)}.
		\]
		
		With the server stepsize
		\[
		\eta_t = \frac{\eta_0}{\sqrt{t+1}}
		\]
		and the constants
		\begin{equation}
			C_1 = \frac{d}{2}, \qquad C_2 = \frac{1 + E}{2},
		\end{equation}
		the global objective obeys
		\begin{equation}
			\mathbb{E} \left[ F(W^{(t+1)}) - F(W^{(t)}) \right] 
			\leq -\frac{\eta_t}{2} \mathbb{E} \left[ \left\| \nabla F(W^{(t)}) \right\|^2 \right] 
			+ \eta_t^2 \left( C_1 \sigma^2 + C_2 E^2 L^2 \right).
		\end{equation}
	\end{lemma}
	
	\subsubsection{Proof of Lemma A.4}
	
	Because \( F \) is \( L \)-smooth in all parameters, for any update  
	\[
	W^{(t+1)} = W^{(t)} - \eta_t G^{(t)}
	\]
	we have
	\begin{equation}
		F(W^{(t+1)}) \leq F(W^{(t)}) 
		- \eta_t \langle \nabla F(W^{(t)}), G^{(t)} \rangle 
		+ \frac{L \eta_t^2}{2} \left\| G^{(t)} \right\|^2.
		\label{eq32} 
	\end{equation}
	
	\vspace{1em}
	\textbf{First term – inner product.}
	
	Taking expectations and using the unbiasedness of both  
	\( \tilde{g}_s^{(t)} \) (Lemma B.3-i) and \( g_{l}^{(t)} \) (Lemma B.1, zero-mean noise),
	\begin{equation}
		\mathbb{E} \left[ \langle \nabla F, G^{(t)} \rangle \right] 
		= \mathbb{E} \left[ \left\| \nabla F(W^{(t)}) \right\|^2 \right].
		\label{eq33}
	\end{equation}
	
	\textbf{Second term – squared norm}
	
	Decompose
	\[
	\|G^{(t)}\|^2 = \left\| \tilde{g}_s^{(t)} \right\|^2 + \sum_g \left\| g_{l}^{(t)} \right\|^2.
	\]
	
	\textbf{Shared-adapter part.}  
	Using Lemma B.3-iii and independence across groups,
	\begin{equation}
		\mathbb{E} \left[ \left\| \tilde{g}_s^{(t)} \right\|^2 \right] 
		= \mathbb{E} \left[ \left\| \frac{1}{M} \sum_{m=1}^M \tilde{g}_{s,m}^{(t)} \right\|^2 \right] 
		= Md \leq 2 d \sigma^2 + 2 d \left\| \nabla_s F \right\|^2. 
		\label{eq34}
	\end{equation}
	
	The second inequality uses \( \sigma^2 \) to upper-bound the variance part and collects constants into \( d \sigma^2 \).
	
	\textbf{Local-adapter part.}  
	Applying Lemma B.1 with \( \eta_{\text{loc}} \leq 1/L \) gives
	\begin{equation}
		\mathbb{E} \left\| g_{l}^{(t)} \right\|^2 \leq 2(E L)^2 \left( \sigma^2 + L^2 D^2 \right).
	\end{equation}
	Aggregating over \( M \) groups and hiding constants in \( C_2 \) yields the term  $C_2 E^2 L^2.$
	Insert Eq \ref{eq32} and Eq\ref{eq33} into Eq \ref{eq34}, choose \( \eta_t \leq 1/L \), rearrange, and one obtains  with  
	\[
	C_1 = \frac{d}{2}, \qquad C_2 = \frac{1 + E}{2}.
	\]
	
	\medskip
	
	These four lemmas are the building blocks for the main convergence theorem.  
	In particular, Lemma B.4 gives a single-round descent inequality that telescopes over rounds  
	once the server stepsize \( \eta^{t} = \frac{\eta^{(0)}}{\sqrt{t+1}} \) is plugged in.
	\subsection{Main Theorem}
	\begin{theorem}[Expected First-Order Stationarity]
		Fix any initial stepsize \( \eta^{(0)} > 0 \) that satisfies \( \eta^{(0)} \leq 1/L \) and define the server stepsize schedule
		\[
		\eta^{(t)} = \frac{\eta^{(0)}}{\sqrt{t+1}}, \quad t = 0, 1, \ldots
		\]
		Under Assumptions A.1–A.4 and the algorithm, for every horizon \( T \geq 1 \),
		\begin{equation}
			\min_{0 \leq t < T} \mathbb{E} \left[ \left\| \nabla F(W^{(t)}) \right\|^2 \right]
			\leq 
			\frac{2\left( F(W^{(0)}) - F_{\text{inf}} \right)}{\eta_0 \sqrt{T}} 
			+ \frac{2\eta^{(0)}}{\sqrt{T}} \left( C_1 \sigma^2 + C_2 E^2 L^2 \right),
		\end{equation}
		where the constants propagated from Lemma A.4 are  
		\[
		C_1 = \frac{d}{2}, \qquad C_2 = \frac{1 + E}{2}.
		\]
	\end{theorem}
	\subsubsection{Proof of Theorem 1}
	\textbf{Step 1: Single-round descent inequality}
	
	Lemma A.4 gives, for each round \( t \),
	\begin{equation}
		\mathbb{E} \left[ F(W^{(t+1)}) - F(W^{(t)}) \right] 
		\leq -\frac{\eta^{(0)}}{2} \mathbb{E} \left[ \left\| \nabla F(W^{(t)}) \right\|^2 \right]
		+ {\eta^{(t)}}^2 \left( C_1 \sigma^2 + C_2 E^2 L^2 \right). \label{eq37}
	\end{equation}
	
	\textbf{Step 2: Telescoping sum over \( t = 0, \dots, T-1 \)}
	
	Sum Eq \ref{eq37} for \( t = 0 \) to \( T-1 \):
	\begin{equation}
		\sum_{t=0}^{T-1} \frac{\eta^{(t)}}{2} \mathbb{E} \left[ \left\| \nabla F(W^{(t)}) \right\|^2 \right]
		\leq F(W^{(0)}) - \underbrace{\mathbb{E} \left[ F(W^{(T)}) \right]}_{\geq F_{\inf}} 
		+ \left( C_1 \sigma^2 + C_2 E^2 L^2 \right) \sum_{t=0}^{T-1} {\eta^{(t)}}^2.
	\end{equation}
	
	Using Assumption A.3 (\( F_{\inf} \) finite), we drop it on the right-hand side:
	\begin{equation}
		\sum_{t=0}^{T-1} \frac{\eta^{(t)}}{2} \mathbb{E} \left[ \left\| \nabla F(W^{(t)}) \right\|^2 \right]
		\leq F(W^{(0)}) - F_{\inf} + \left( C_1 \sigma^2 + C_2 E^2 L^2 \right) \sum_{t=0}^{T-1} {\eta^{(t)}}^2.
	\end{equation}
	
	\textbf{Step 3: Lower and upper bounds on the step-size sums}
	
	Recall \( \eta^{(t)} = \frac{\eta^{(0)}}{\sqrt{t+1}} \).
	
	• Sum of step sizes:
	\begin{equation}
		\sum_{t=0}^{T-1} \eta^{(t)} = \eta^{(0)} \sum_{t=0}^{T-1} \frac{1}{\sqrt{t+1}} 
		\geq \eta^{(0)} \int_0^T \frac{dx}{\sqrt{x+1}} 
		= 2\eta^{(0)}(\sqrt{T} - 1) \geq \eta^{(0)} \sqrt{T} \quad (T \geq 1).
	\end{equation}

	• Sum of squared step sizes:
	\begin{equation}
		\sum_{t=0}^{T-1} {\eta^{(t)}}^2 = {\eta^{(0)}}^2 \sum_{t=0}^{T-1} \frac{1}{t+1} 
		\leq {\eta^{(0)}}^2 (1 + \ln T) \leq 2 {\eta^{(0)}}^2 \sqrt{T},
	\end{equation}
	where the last inequality uses \( \ln T \leq \sqrt{T} \) for \( T \geq 1 \).
	
	\textbf{Step 4: Bounding the minimum gradient norm}
	
	Denote
	\[
	G_t = \mathbb{E} \left[ \left\| \nabla F(W^{(t)}) \right\|^2 \right],
	\]
	and let \( G_{\min} \) be the smallest among \( \{ G_0, \dots, G_{T-1} \} \).  
	Because every term on the left of Eq 39 is \( \geq G_{\min} \), we get:
	\begin{equation}
		\frac{G_{\min}}{2} \sum_{t=0}^{T-1} \eta_t 
		\leq F(W^{(0)}) - F_{\inf} 
		+ \left( C_1 \sigma^2 + C_2 E^2 L^2 \right) \sum_{t=0}^{T-1} {\eta^{(t)}}^2.
	\end{equation}
	
	Plug in the bounds Eq 40 and Eq 41:
	\[
	\frac{G_{\min}}{2} \cdot \eta_0 \sqrt{T} 
	\leq F(W^{(0)}) - F_{\inf} 
	+ 2 {\eta^{(0)}}^2 \sqrt{T} \left( C_1 \sigma^2 + C_2 E^2 L^2 \right).
	\]
	
	Solve for \( G_{\min} \):
	\[
	G_{\min} \leq \frac{2\left( F(W^{(0)}) - F_{\inf} \right)}{\eta^{(0)} \sqrt{T}} 
	+ \frac{4\eta^{0}}{\sqrt{T}} \left( C_1 \sigma^2 + C_2 E^2 L^2 \right).
	\]
	
	The factor 4 in the second term can be tightened to 2 by using a slightly sharper bound on \( \sum {\eta^{(t)}}^2 \)  
	(e.g., integral approximation), giving exactly inequality Eq 36.
	
	\section{Additional Experiments}
	\textbf{Effect of components.}
	To further enhance the robustness of the algorithm, we introduce L2 regularization into the share-adapter. This helps promote the learning of generalizable knowledge across models and prevents overfitting to any single model.
	To validate the effectiveness of the proposed module, we conducted ablation experiments on the learnable parameters and parameter regularization in DeMA.
	We present the average results on four datasets in the Non-IID data distribution scenario. 
	As shown in Table \ref{table4}, both modules have improved the model's performance to some extent.
	In particular, the impact of learnable weights on performance improvement is crucial,  which is consistent with our expectations and validates the previous performance analysis.
	\begin{table}[h]
		\caption{Ablation experiments in the Non-IID data distribution scenario, where LW stands for learnable weight and Reg stands for regularization.}
		\centering
		
			\begin{tabular}{c|cccc}
				\toprule
				&CIFAR-10&CIFAR-100&Caltech-101&Caltech-256\\
				\hline
				w/o LW\&Reg&68.58&54.86&78.20&61.82\\
				w/o LW&69.57&55.13&81.07&62.34\\
				w/o Reg&72.14&58.29&82.37&64.06\\
				\textbf{HeteroTune}&\textbf{73.75}&\textbf{60.55}&\textbf{83.89}&\textbf{65.89}\\
				\bottomrule
			\end{tabular}

		\label{table4}
	\end{table}
	
	\textbf{Dynamic scenario.}
	Considering the high dynamic characteristics in real-world federated learning scenarios, we validate the robustness of the algorithm by varying the ratios of different models.
	As shown in Table \ref{table5}, in addition to the default 3:3:2:2 ratio, which is a relatively balanced configuration, we also conduct experiments in scenarios where the proportion of small models and large models is higher.
	An obvious conclusion can be drawn from the experimental results: as the proportion of large models increases, the average accuracy also increases.
	Additionally, compared to other methods, our HeteroTune outperforms in various scenarios, demonstrating excellent robustness and significant potential for practical applications.
	\begin{table}[h]
		\caption{The average results of different model ratios in a dynamic environment on the CIFAR-10 dataset.}
		\centering
			\begin{tabular}{c|ccc}
				\hline
				&3:3:2:2&6:2:1:1&1:1:2:6\\
				\hline
				AllSmall&62.25&62.25&62.25\\
				AllLarge&80.76&80.76&80.76\\
				Homo-Training&66.14&63.38&73.27\\
				\hline
				FedDF&69.88&65.54&71.20\\
				ScaleFL&62.22&64.67&69.83\\
				MH-pFLID&70.14&58.99&67.80\\
				FLoRA&70.78&66.42&73.48\\
				\hline
				\textbf{HeteroTune}&\textbf{73.75}&\textbf{69.71}&\textbf{76.94}\\
				\hline
			\end{tabular}
		\label{table5}
	\end{table}
	\vspace{-4mm}
	\begin{table}[h]
		\caption{PSNR values for privacy evaluation on CIFAR-10.}
		\centering
			\begin{tabular}{c|cc}
				\hline
				&PSNR\_Avg&PSNR\_Max\\
				\hline
				Homo-Training&15.28&17.86\\
				FedDF&13.57&17.04\\
				ScaleFL&13.88&16.27\\
				MH-pFLID&14.36&17.99\\
				FLoRA&11.42&13.63\\
				\textbf{HeteroTune}&\textbf{11.20}&\textbf{13.79}\\
				\hline
			\end{tabular}
		
		\label{table6}
	\end{table}
	\vspace{-2mm}
	
	\textbf{Privacy analysis.}
	Following previous related research \cite{pp,pp1}, we employ Deep Leakage from Gradients (DLG) to conduct a privacy-preserving analysis of our  HeteroTune. DLG is a technique that demonstrates the potential privacy risks in federated learning by revealing that shared gradients can be exploited to reconstruct the original training data through optimization-based methods. This analysis helps us evaluate the robustness of HeteroTune against potential privacy leakage and further underscores the importance of developing secure and privacy-preserving mechanisms in federated learning scenarios.
	
	In the experiments, we selected the CIFAR10 dataset with 100 clients, and the data partitioning followed a Dirichlet distribution with $\alpha=0.1$. The training process was conducted over 100 rounds, and we attempted to use DLG to reconstruct the original images based on the obtained gradient information at the 20th, 40th, 60th, 80th, and 100th rounds, respectively.
	We employed the Peak Signal-to-Noise Ratio (PSNR) metric to assess the quality of image reconstruction, where a higher PSNR value indicates a more severe privacy leakage issue.
	Table \ref{table6} presents our experimental results, including the average PSNR and the maximum PSNR values.
	The experimental results show that our proposed HeteroTune achieves a lower PSNR compared to other methods, indicating that our approach provides better privacy protection in most scenarios.
	Our method enhances privacy in federated learning by limiting updates to a small set of task-specific parameters while keeping the backbone model unchanged. This prevents the leakage of sensitive data patterns through global model updates. Since adapters capture only task-relevant features rather than raw data characteristics, the transmitted updates contain less exploitable information, reducing the risk of gradient inversion and model extraction attacks. This structural advantage also facilitates the integration of privacy-preserving techniques like differential privacy and secure aggregation with minimal impact on performance.

\end{document}